\documentclass[sigconf]{acmart}

\copyrightyear{2021}
\acmYear{2021}
\setcopyright{acmcopyright}
\acmConference[K-CAP '21] {Proceedings of the 11th Knowledge Capture Conference}{December 2--3, 2021}{Virtual Event, USA.}
\acmBooktitle{Proceedings of the 11th Knowledge Capture Conference (K-CAP '21), December 2--3, 2021, Virtual Event, USA}
\acmPrice{15.00}
\acmISBN{978-1-4503-8457-5/21/12}
\acmDOI{10.1145/3460210.3493566}

\usepackage[utf8]{inputenc}
\usepackage{natbib}
\usepackage{graphicx}
\usepackage{caption}
\usepackage{subcaption}
\usepackage{mathtools}
\usepackage{multirow}
\usepackage{booktabs}
\usepackage{tablefootnote}
\usepackage{threeparttable}
\usepackage{float}
\usepackage{appendix}
\usepackage{comment}

\usepackage{hyperref}

\newcommand{\rulesep}{\unskip\ \vrule\ }

\begin{document}
\fancyhead{}

\title{Time in a Box: Advancing Knowledge Graph Completion with Temporal Scopes}

\author{Ling Cai}
\email{lingcai@ucsb.edu}
\affiliation{%
  \institution{UC Santa Barbara}
}
\author{Krzysztof Janowicz}
\affiliation{%
  \institution{UC Santa Barbara}
}
\author{Bo Yan}
\affiliation{%
  \institution{UC Santa Barbara}
}
\author{Rui Zhu}
\affiliation{%
  \institution{UC Santa Barbara}
}

\author{Gengchen Mai}
\affiliation{%
  \institution{Stanford University}
}

\renewcommand{\shortauthors}{Cai and Janowicz, et al.}

\begin{abstract}
Almost all statements in knowledge bases have a temporal scope during which they are valid. Hence, knowledge base completion (KBC) on temporal knowledge bases (TKB), where each statement \textit{may} be associated with a temporal scope, has attracted growing attention. Prior works assume that each statement in a TKB \textit{must} be associated with a temporal scope. This ignores the fact that the scoping information is commonly missing in a KB. Thus prior work is typically incapable of handling generic use cases where a TKB is composed of temporal statements with/without a known temporal scope. In order to address this issue, we establish a new knowledge base embedding framework, called TIME2BOX, that can deal with atemporal and temporal statements of different types simultaneously. Our main insight is that answers to a temporal query always belong to a subset of answers to a time-agnostic counterpart. Put differently, time is a filter that helps pick out answers to be correct during certain periods. We introduce boxes to represent a set of answer entities to a time-agnostic query. The filtering functionality of time is modeled by intersections over these boxes. In addition, we generalize current evaluation protocols on time interval prediction. We describe experiments on two datasets and show that the proposed method outperforms  state-of-the-art (SOTA) methods on both link prediction and time prediction.
\end{abstract}

\ccsdesc[500]{Computing methodologies~Knowledge representation and reasoning}
\ccsdesc[500]{Computing methodologies~Temporal reasoning}

\keywords{Temporal Knowledge Base, TIME2BOX, Link Prediction, Time Prediction}
\maketitle

\section{Introduction} \label{sec:intro}
A knowledge base (KB) such as Wikidata and DBpedia stores statements about the world around us. A KB is typically represented as a set of triples in the form of $(s, r, o)$ -- short for \textit{(subject, relation, object)}, encoding the association between entities and relations among them. A statement is often temporally scoped, which indicates during which time period it is valid. Two examples are (\textit{Albert Einstein, educatedAt, ETH Zurich, 1896 - 1900}) and (\textit{Albert Einstein, academicDegree, Doctor of Philosophy in Physics, 1906}). The former specifies the time period during which Albert Einstein studied at ETH, and the latter points out the specific date when he obtained his  degree. Graphs that contain a substantial amount of such time-aware statements are often called temporal knowledge base (TKB) in the machine learning literature. Each statement in a TKB is associated with a validity time as (\textit{s, r, o, $t^*$})\footnote{$t^{*}$ could be a time instant or time interval}.

Due to the ever-changing state of the world and missing data, TKBs usually contain inaccurate and incomplete information similar to KBs. The sparsity of TKBs necessitates temporal knowledge base completion (TKBC), namely inferring missing statements from known statements. Temporal link prediction task is proposed to evaluate a TKBC model by testing its performance on answering incomplete temporal queries of the form (\textit{s, r, ?o, $t^*$}) or  (\textit{?s, r, o, $t^*$}).



Despite recent success stories on time-agnostic KBC, research on TKBC is still in its early age and is facing  new challenges. The validity time period of a statement is often missing in a KB. As a result, it is difficult to distinguish whether statements in a KB are atemporal (e.g., (\textit{Albert Einstein,~instanceOf,~Human})) or time-dependent (e.g., (\textit{United States of America, instanceOf, Historical Unrecognized State\footnote{According to Wikidata that statement holds true during 1776-1784.}})). This leads to the question of which statements should be part of a TKB in the first place. Prior works restrict TKBs to a collection of statements where the validity time period for each statement \textit{must} be available. However, in WIKIDATA114k, a dataset from Wikidata, for instance, 85.1\% of all statements are temporal, 56.2\% of the temporal statements are missing their validity time information and are excluded in previous studies while only 247,393 out of 1,660,824 statements (i.e., 14.9\%) are truly atemporal
\footnote{For all the statements, we first categorize predicates into two groups -- atemporal predicates and temporal predicates. If a predicate has ever been involved in a statement that has temporal scoping, it belongs to temporal predicates; otherwise, it is an atemporal predicate. Atemporal statements are those associated with atemporal predicates.}. As the number of temporal statements with missing validity information is substantial, excluding them from a TKB will significantly reduce the amount of information that could be useful in TKB studies.
    
Retaining these temporally scoped statements leads to several challenges that need to be addressed. For instance, how to design a TKBC model to handle statements with and without known temporal scoping from the data representation perspective and model design perspective? Clearly, the conventional representation in prior TKBC in the form of (\textit{s, r, o, t})\footnote{t denotes a time point} falls short. An ideal TKBC model should be more \textit{flexible} to address cases when the validity information of different types (i.e., point in time, right-open interval (known start time), left-open interval (known end time), closed interval) 
is presented in a TKB or even no validity information is available for a statement. 

The second challenge is how to predict the temporal scope of a statement as it is often missing in TKBs. This task is referred to as time interval prediction, which amounts to answering incomplete queries of the form (\textit{s, r, o, $?I$}). How to generate a predicted time interval and evaluate it require further investigation. This problem has only been addressed very recently by ~\citet{jain-etal-2020-temporal}. However, at times their evaluation protocols fail to distinguish one predicted interval from another since they do not consider the gap between the predicted and the gold interval in case of no overlap. For instance, the same metric scores are assigned to two predictions [1998, 1999] and [1998, 2010] when a gold interval [2011, 2020] is considered. 

\begin{figure*}
    \centering
    \begin{subfigure}[b]{0.33\textwidth}
        \centering
        \includegraphics[width=\textwidth]{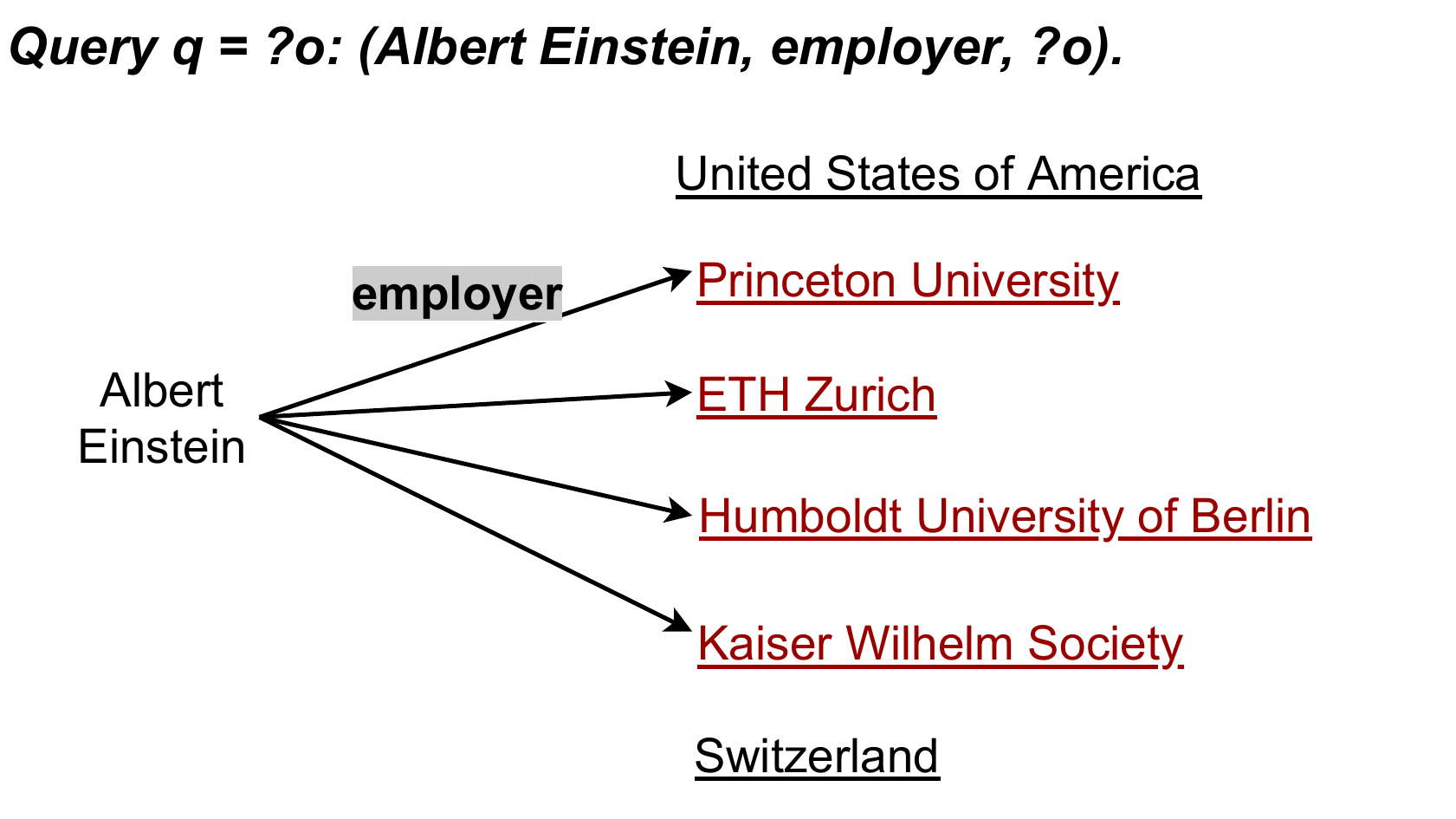}
        \label{fig:atemporal_kb}
    \end{subfigure}
    \rulesep
     \begin{subfigure}[b]{0.32\textwidth}
        \centering
        \includegraphics[width=\textwidth]{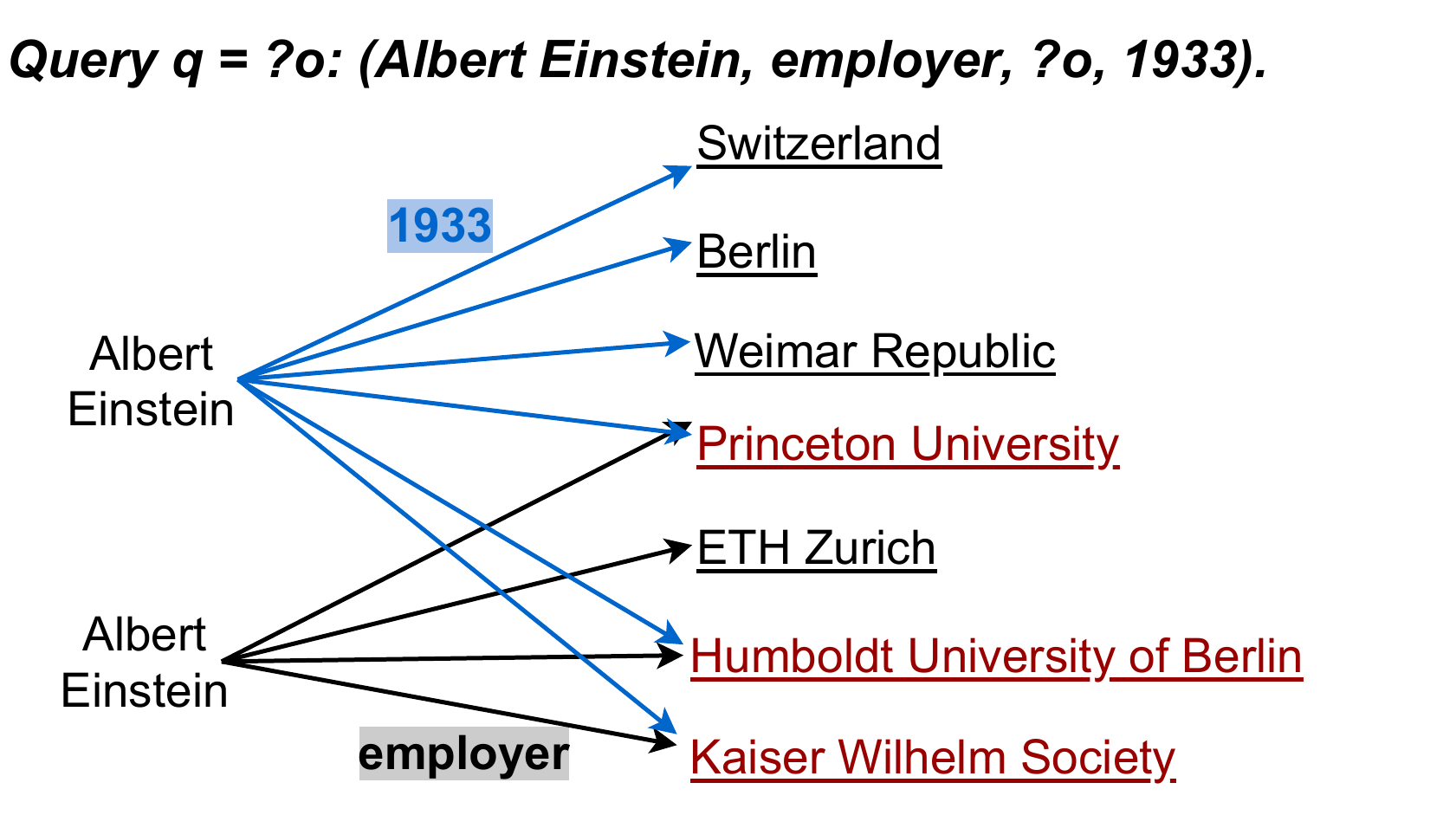}
        \label{fig:instant_kb}
    \end{subfigure}
      \rulesep
        \begin{subfigure}[b]{0.33\textwidth}
        \centering
        \includegraphics[width=\textwidth]{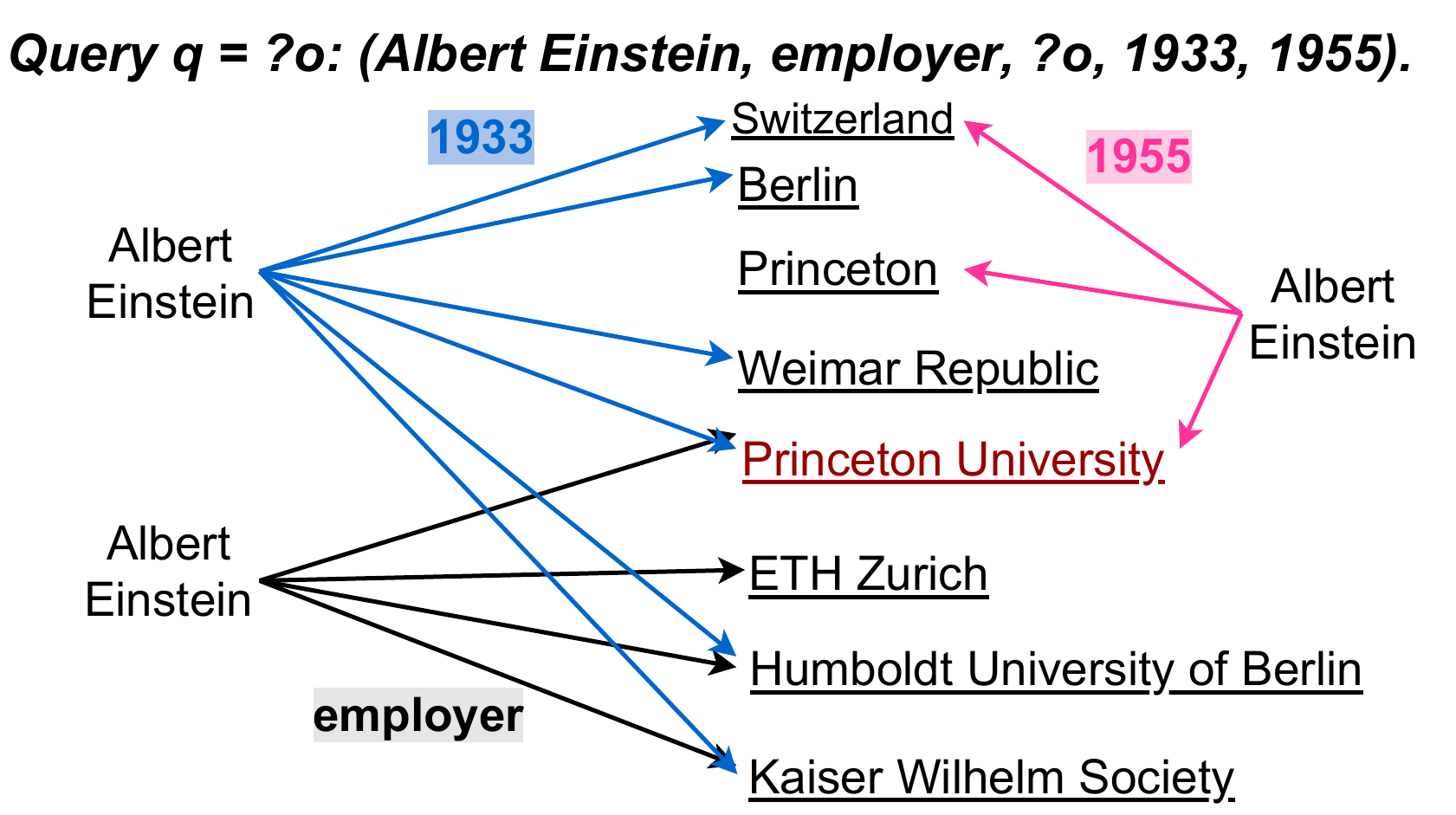}
        \label{fig:interval_kb}
    \end{subfigure}
    \begin{subfigure}[b]{0.32\textwidth}
        \centering
        \includegraphics[width=1.02\textwidth]{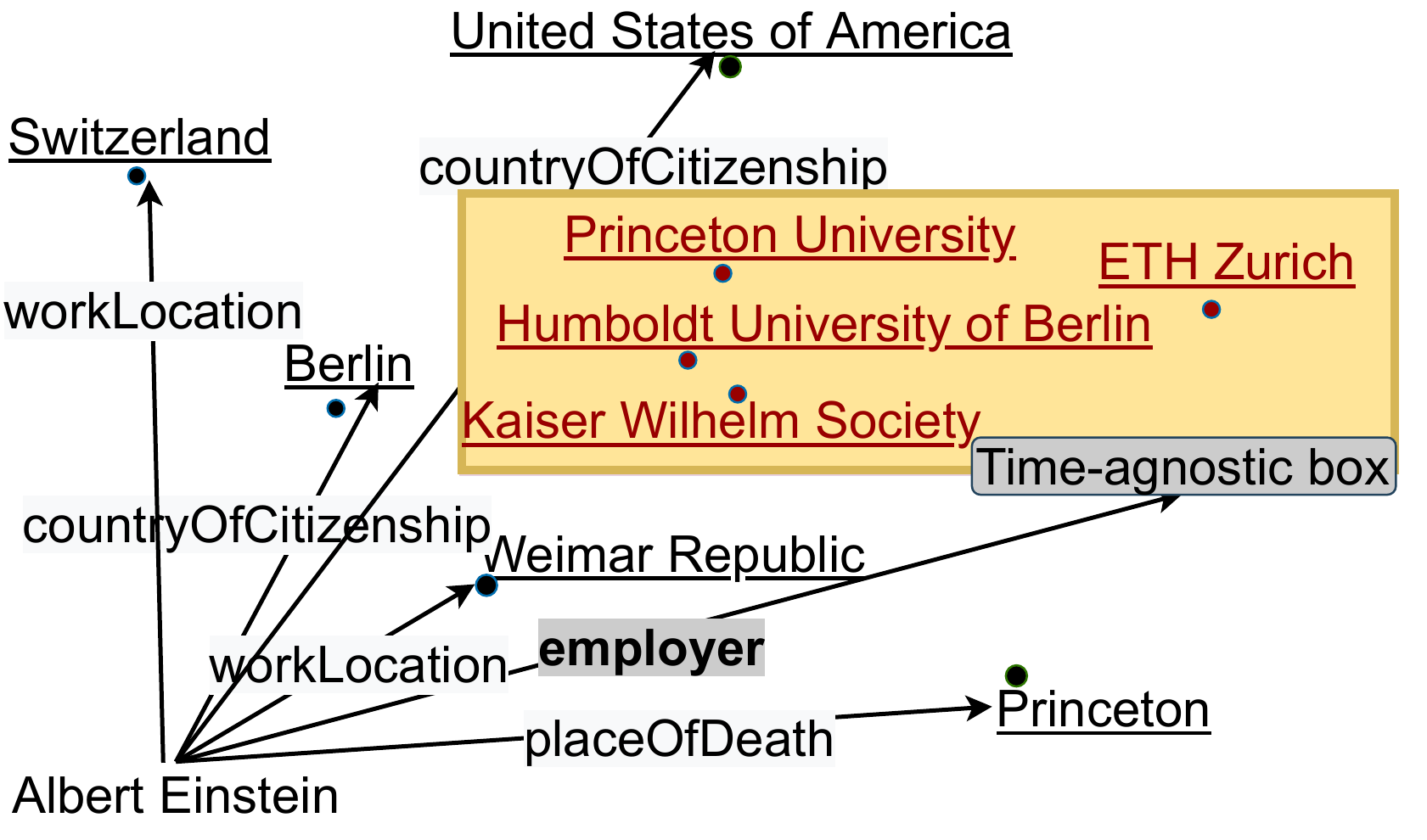}
        \caption{Modeling an atemporal statement}
        \label{fig:atemporal_vs}
    \end{subfigure}
      \rulesep
    \begin{subfigure}[b]{0.32\textwidth}
        \centering
        \includegraphics[width=1.0\textwidth]{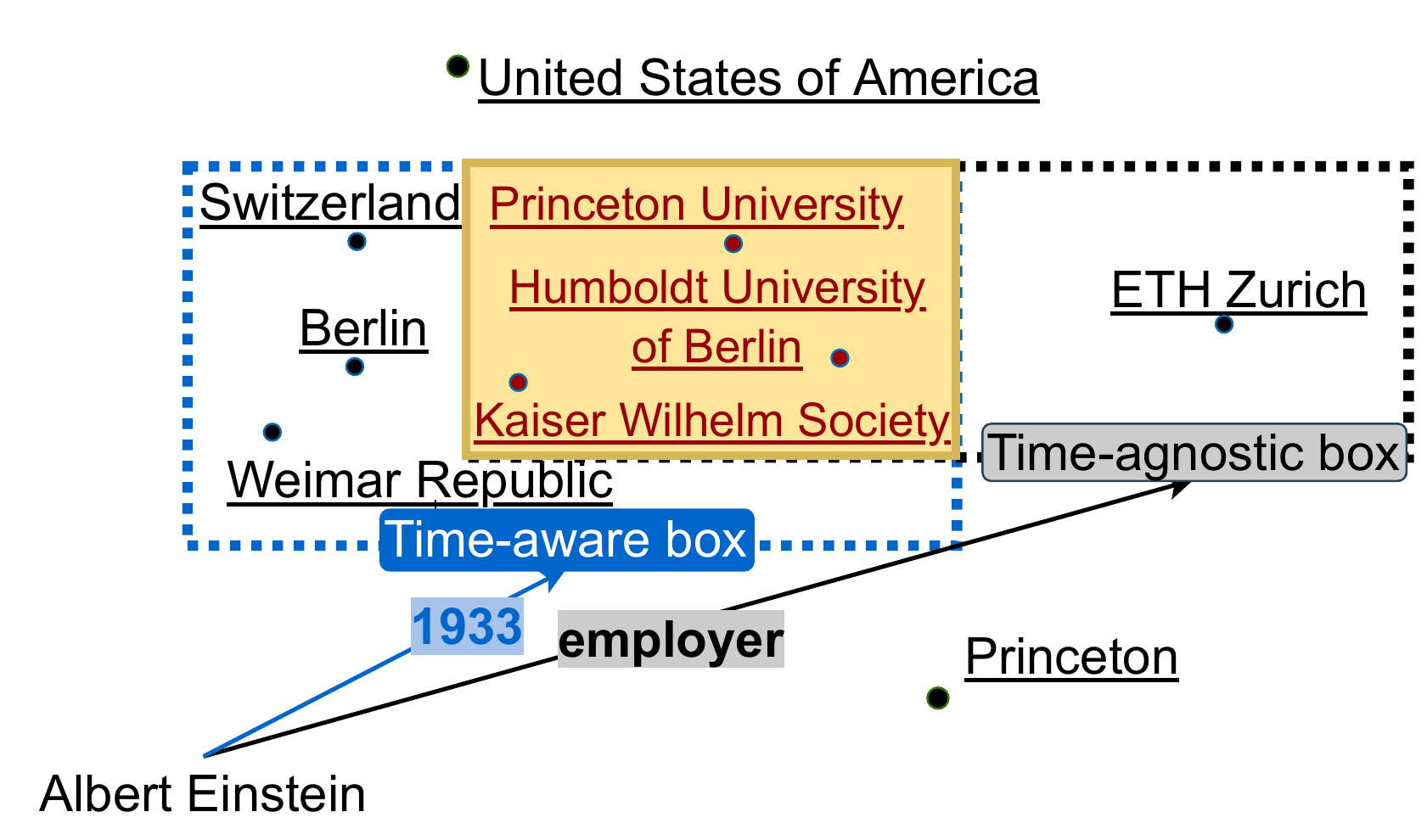}
        \caption{Modeling a timestamp-based statement}
        \label{fig:instant_vs}
    \end{subfigure}
      \rulesep
    \begin{subfigure}[b]{0.32\textwidth}
        \centering
        \includegraphics[width=\textwidth]{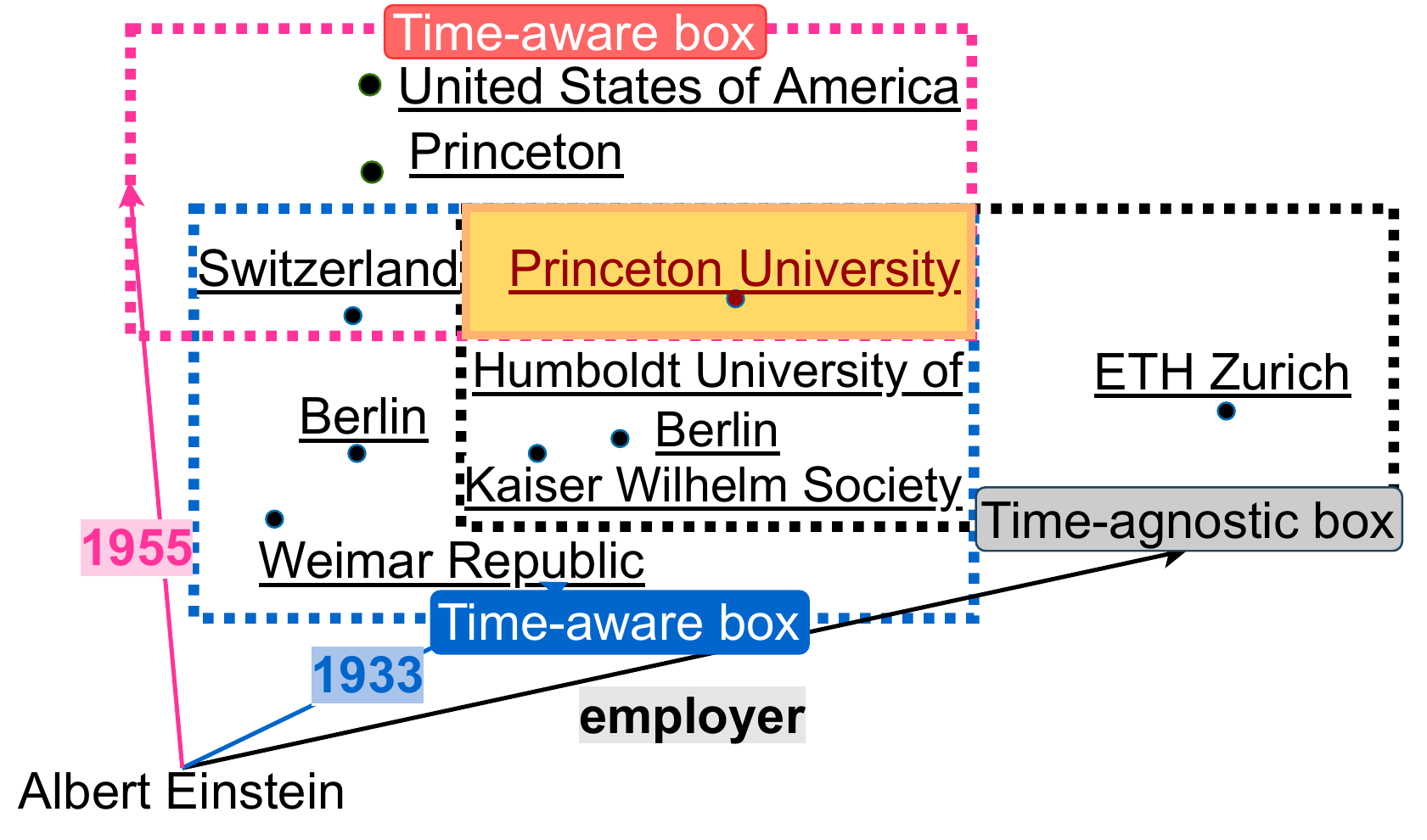}
        \caption{Modeling an interval-based statement}
        \label{fig:interval_vs}
    \end{subfigure}

        \caption{\textbf{Illustration of TIME2BOX reasoning process.} In each figure, the upper part shows entities and relations in the KB space and the latter illustrates their correspondences in the embedding space. In all figures, the final boxes are shaded regions in orange and answer entities are in the boxes. Note that we omit the edges between \textit{Albert Einstein} and associated entities in Fig.\ref{fig:instant_vs} and \ref{fig:interval_vs} for simplicity. Fig.\ref{fig:atemporal_vs} shows that for an atemporal query, the reasoning process picks out all possible answers from the whole entity space and encloses them into a time-agnostic box. In Fig.\ref{fig:instant_vs} a time-aware box is added to enclose entities that are relevant to \textit{Albert Einstein} in 1933. Then the intersection between time-agnostic and time-aware boxes consists of a new box, which contains entities that satisfy both requirements. When more validity information is available, Fig.\ref{fig:interval_vs} shows that more time-aware boxes can be added and the intersection box that contains correct answers would shrink further. By doing so, TIME2BOX is flexible enough to handle different types of queries.}
    \label{fig:time_reasoning_process}
\end{figure*}

In this paper, we present a novel TKBC embedding framework, called TIME2BOX, which relies on the intuition that the answer set in a temporal query $(s, r, ?o, t*)$ is always  a subset of answers of its time-agnostic counterpart $(s, r, ?o)$. As illustrated in Fig.~\ref{fig:time_reasoning_process}, there are four correct answer entities to a query (\textit{Albert Einstein, employer, ?o}). However, when time information is specified (e.g., \textit{Albert Einstein, employer, ?o, 1933}) as shown in Fig. \ref{fig:instant_vs}, the number of positive answers becomes three. With more temporal information being available (e.g., \textit{Albert Einstein, employer, ?o, [1933, 1955]}), the answer set shrinks further (see Fig.~\ref{fig:interval_vs}). Therefore, we propose to model a statement in a TKB by imitating the process of answering its corresponding temporal query$(s, r, ?o, t^{*})$, which can be achieved in two steps -- finding answer entities to its atemporal counterpart (\textit{s, r, ?o}) by using KBC methods and then picking out entities to be true to the temporal query from preceding answers by including time. We implement this idea by using box embeddings~\citep{vilnis2018probabilistic, patel2020representing}, especially inspired by QUERY2BOX~\citep{ren2019query2box}, which is originally used for answering conjunctive queries~\citep{mai2019contextual}. 
Boxes, as containers, can naturally model a set of answers they enclose. The filtering functionality of time can be naturally modeled as intersections over boxes similarly to Venn diagrams~\citep{venn1880diagrammatic}. Meanwhile, performing an intersection operation over boxes would still result in boxes, thus making it possible to design a unified framework to deal with statements/queries of different types.

\textbf{Our main research contributions are listed as follows}:

\begin{itemize}
\item We propose a box-based knowledge graph embedding TKBC framework (TIME2BOX) that can represent and model statements with different types of validity information (i.e., unknown temporal scoping, left/right-open intervals and closed intervals).
\item  We introduce a new evaluation metrics $gaeIOU$ for interval evaluation by taking the gap between a gold interval and a predicted interval into consideration if no overlap between them exists.
\item Extensive experiments on two datasets - WIKIDATA12k and WIKIDATA114k - show that TIME2BOX yields the state-of-the-art (SOTA) results in link prediction and outperforms SOTAs in time prediction by significant margins. 
\end{itemize}

TIME2BOX code is available in Github\footnote{\url{https://github.com/ling-cai/Time2Box}}.

\section{Related Work} \label{related_work}
\paragraph{Knowledge Base Completion}
KBC has been extensively studied in the past~\citep{bordes2013translating, lin2015modeling, yang2014embedding, trouillon2016complex, sun2019rotate}. The core insight of these methods is to embed entities and relations in a KB into low-dimensional vectors, which can be utilized in downstream tasks, such as link prediction. These methods can be roughly classified into two groups: transformation-based models and semantic matching energy based models. Transformation-based models treat a relation as a transformation operator. Two well-known assumptions are translation (e.g., TransE~\citep{bordes2013translating}) and rotation (e.g., RotatE~\citep{sun2019rotate}). For instance, TransE assumes that for a statement (\textit{s, r, o}), the object embedding can be derived by translating the subject embedding operated by the relation embedding in the embedding space. As such, the presence of a statement in a KG is measured by the distance between the object embedding and the subject embedding after transformation. Semantic matching energy based methods determine the existence of a statement by a score calculated from a function of learned entity and relation embeddings in the latent space~\citep{yang2014embedding, trouillon2016complex}. For instance, DistMult~\citep{yang2014embedding} uses a 3-way inner product as the scoring function. In addition to these basic triple-based methods, other studies have focused on exploiting higher-order structural information in a KG (e.g., paths, neighbours), such as PTransE~\citep{lin2015modeling}, R-GCN~\citep{schlichtkrull2018modeling}, and TransGCN~\citep{cai2019trans}. All KBC models ignore the temporal scoping of statements, 
and thus are unable to address temporal statements. However, these models are the foundations for TKBC.

\paragraph{Temporal Knowledge Base Completion}
Recently, there has been a surge of interest in taking validity information into consideration as KB statements are usually time-dependent. There are two lines of works on temporal link prediction. The first branch focuses on so-called dynamic knowledge bases (i.e., event KBs (ICEWS)~\cite{DVN/28075_2015}), where each statement is associated with a timestamp. The insight behind this branch is that knowledge in KBs evolves over time and historical statements/events drive the occurrence of new events. Therefore, their focus is more on extrapolation -- predicting unseen entity relationships over time by modeling temporal dependencies of statements/events in KBs ~\citep{trivedi2017know, xu2019temporal, jin2020recurrent, deng2020dynamic}. The most well-known model is Know-Evolve~\citep{trivedi2017know}, which assumes that the occurrence of facts/events can be modeled as a multivariate temporal point process. 

Unlike the first branch that assumes timestamp-based statements/facts, a TKB in the second branch can associate a statement with time instants or time intervals as its validity information. Moreover, the goal of this line of work is more about interpolation -- filling in missing components in TKGs with/without explicitly modeling the temporal dependencies between statements. Recent works follow a common paradigm, that is, to encode time as embeddings and then incorporate them into time-agnostic KBC models~\citep{Leblay2018deriving, garcia-duran-etal-2018-learning, goel2020diachronic, ma2019embedding, lacroix2019tensor, jain-etal-2020-temporal}. \citet{Leblay2018deriving} investigated several extensions of existing KBC models by directly fusing time embeddings with relation embeddings, including TTransE, TRESCAL, etc. \citet{goel2020diachronic} proposed to learn time-varying entity embeddings by replacing a fraction of embedding weights with an activation function of learned frequencies. Unlike previous work, which view time as numeric values, \citet{garcia-duran-etal-2018-learning} concatenated the string representation of the relation and the time, and fed them into an LSTM to obtain time-aware relation representations, which were used in TransE (TA-TransE) and DistMult(TA-DM) afterwards. More recently, \citet{lacroix2019tensor} presented the TKBC problem as a four-way tensor completion problem, and proposed TNTComplEx, which was extended from the time-agnostic ComplEx~\cite{trouillon2016complex}. \citet{jain-etal-2020-temporal} augmented TNTComplex with three more time-dependent terms as analogy to the idea of approximating joint distributions by low-order marginals in graph models and incorporated soft ordering and span constraints as temporal inductive biases. Our work belongs to the latter group. However, our proposal is more flexible as we can deal with cases when $t^*$ is a time instant, a (left/right-open) interval, closed interval or even missing, while prior works can only handle one timestamped representation of the form ($s$, $r$, $o$, $t$).

\section{Preliminaries} \label{sec:preliminary}
\subsection{Temporal Knowledge Bases}
Prior TKBC methods typically work on TKBs in which each statement has to be associated with validity information. Thereby, for statements that do not have known temporal scopes, they either exclude them from a TKB in the beginning or assume that these statements hold all the time~\citep{lacroix2019tensor}. However, there are limitations in both ways. As discussed in Section~\ref{sec:intro}, excluding them from a TKB will significantly reduce the amount of information that could be beneficial in TKBC studies as the number of such statements is substantial. For the latter, their assumption would be problematic since a lot of them may only hold for a certain time period. For instance, the statement (\textit{Warsaw, country, Russian Empire}) holds during the time interval [1815-07-09, 1916-11-04]. Following the open-world assumption (OWA), we argue that TKBs are an extension to KBs insofar as the lack of temporal scoping for any given statement does not imply it holding indefinitely.

In the following, we use $t$ and $I_{st}^{et}=[st, et]$ to denote a time point and a time interval, respectively. The symbol $-$ will stand for unknown temporal validity. There are five types of statements in such a TKB: (1) ({$s$, $r$, $o$}) for a statement without a known temporal scope; (2) ($s$, $r$, $o$, $t$) for a timestamped statement which holds at a point in time $t$; (3) ($s$, $r$, $o$, $I^-_{st}$) for a right-open interval-based statement, in which only the time when the statement starts to hold is known; (4) (\textit{s, r, o, $I^{et}_{-}$}) for a left-open interval-based statement, in which only the time when the statement ceases to hold is known; and (5) ({$s$, $r$, $o$, $I^{et}_{st}$}) for a statement which is temporally scoped by a closed interval $I_{st}^{et}$. Then a TKB is denoted as $\mathcal{G}=\bigcup_{(s,r,o,t^*)}$, namely the union  of statements of the five types, where $s,o \in E$ represent entities, $r \in R$ denotes a relation and $t^* \in \{t, I_{st}^{-}, I_{-}^{et}, I_{st}^{et}, None\}$ denotes different types of valid time or no valid time available. 
 
\subsection{The TKBC Problem}
Link prediction and time prediction are two main tasks used to evaluate a TKBC model. Statements in TKBs are split into training, validation, and test sets, used for model training, parameter tuning and model evaluation, respectively. 
\paragraph{Link prediction} Queries used in this task are of the form ({$s$, $r$, $?o$, $t^{*}$}). Performance is evaluated on the rank of a given golden, i.e., ground truth, answer in the list of all the entities sorted by scores in a descending order. Then MRR (mean reciprocal rank), MR (mean rank), HITS@1, HITS@3 and HITS@10 are computed from the ranks over all queries in the test set. However a query may be satisfied by multiple answer entities. Thus another correct answer may be ranked over the given golden answer. In such cases, a KBC/TKBC model should not be penalized. A traditional strategy used in KBC is to filter out those correct answers that are already in the training and validation sets before calculating metrics. This strategy can be directly applied to queries of the form ({$s$, $r$, $?o$}) or ({$s$, $r$, $?o$, $t$}). However, it may not be sufficient for queries of the form (\textit{s, r, ?o, I}), as there may exist other answers that are true during a time period within the interval~$I$. For example, suppose two statements -- \textit{(Albert Einstein, employer, Princeton University, [1933, 1955])} and \textit{(Albert Einstein, employer, Leiden University, [1920, 1946])}, both Princeton University and Leiden University are correct answers during the period [1933, 1946]. One naive way to solve this problem is to discretize the interval $I$ to a sequence of time points $t$s  and then to convert (\textit{s, r, ?o, I}) into timestamped queries of the form (\textit{s, r, ?o, t}) so that the same filtering process can be performed on each timestamped query. Finally, the ranks over them are averaged to be the rank for a time interval-based query. This idea is well-aligned with the proposal by~\citep{jain-etal-2020-temporal}.
\paragraph{Time prediction} Time prediction queries in TKBs are of the form (\textit{s, r, o, ?I}). Despite the fact that the validity information could be a point in time or a time interval, a point in time can be viewed as a special time interval, in which start time and end time coincide. 
Thus, time prediction boils down to time interval prediction. Its performance is evaluated by the overlap between a gold interval and a predicted interval or the closeness between those in case of no overlap.
We describe the existing evaluation protocols and propose a generalized evaluation metric in Section~\ref{sec:eval_metrics}. 

\section{Method} \label{sec:method}
The key insight of TIME2BOX lies in an intuition that 
the answer set of a temporal query ($s$, $r$, $?o$, $t^*$) is always a subset of answers of its time-agnostic counterpart ($s$, $r$, $?o$) and set size decreases by adding more temporal constraints.
As illustrated in Fig.~\ref{fig:atemporal_vs}, four object entities 
satisfy the atemporal query (\textit{Albert Einstein, employer, ?o}) while three entities are the correct answers when the query is restricted to the year of 1993 (see Fig.~\ref{fig:instant_vs}) and only one entity is correct when another time information is further added in the statement, shown in Fig.\ref{fig:interval_vs}. Inspired by this observation, we propose to model a temporal statement ($s$, $r$, $o$, $t^*$) by imitating the process of answering its corresponding temporal query ($s$, $r$, $?o$, $t^*$),
which can be achieved through two steps: 1) finding a set of answer entities that are true for the corresponding atemporal query by using any KBC model and 2) imposing a filtering operation enforced by time to restrict answers afterwards. In following sections, we take a time instant-based statement as example to formalize our idea in a KB space and a vector space, respectively.

\subsection{Formalization in a KB Space}
For a statement (\textit{s, r, o, t}) in a TKB, the first step of TIME2BOX, as shown in Fig.~\ref{fig:atemporal_vs}, is to project the subject $s$ to a set of object entities that are true to its corresponding atemporal query in the form of (\textit{s, r, ?o}) enforced by the relation $r$. This is a prerequisite for any statement and can theoretically be addressed by any KBC method. Formally, the relation projector is defined as:

\textbf{Relation Projector -- $\mathbf{OP_r}$:} Given the subject entity $s$ and the relation $r$, this operator obtains: $S_r=\{o^{\prime} \mid (s, r, o^{\prime}) \in \mathcal{G}^N\}$. $\mathcal{G}^N$ is the time-agnostic counterpart of $\mathcal{G}$.

Then time information is used to filter out entities that are incorrect during the time of interest from the answer set $S_r$. This can be achieved by first \textit{projecting} the subject $s$ to a set of object entities that co-occur with $s$ in statements at a given time point (as shown in blue edges in Fig.~\ref{fig:instant_vs}) and then finding the \textit{intersection} over them and  $S_r$ (see the three entities in red in Fig.~\ref{fig:instant_vs}). Accordingly, the two involved steps are defined as:

\textbf{Time Projector -- $\mathbf{OP_t}$:} Given the subject $s$ and the timestamp $t$, this operator obtains: $S_t = S_t = \{o^\prime \mid o^\prime \in E ~and ~(s, r^{\prime}, o^{\prime}, t) \in \mathcal{G} ~and~ r^{\prime} \in R \}$.

\textbf{Intersection Operator -- $\mathbf{OI}$:} Given $S_r$ and $S_t$, this operator obtains the intersection $S_{inter}=\{o \mid o \in (S_r ~and~ S_t)\}$.

In fact, such a modeling process also fits to left/right-open interval-based statements directly. For a left/right-open interval-based statement, we only consider the known endpoint time in such an interval as we follow the open-world assumption. However, for an atemporal statement, we only need one relation projector to obtain $S_{r}$, which is the final set consisting of correct answer entities to its query form. For a closed interval-based query, one commonly used approach is to randomly pick one timestamp within the interval and to associate it with (\textit{s, r, o}). Then it can be modeled the same way as an instant-based statement. At training, a timestamp is always randomly picked from the interval to ensure that all the timestamps in the interval are used. In addition to the common strategy, TIME2BOX allows sampling of a sub-time interval within the given interval so that two time constraints (i.e., start time and end time) can be imposed by using two time projectors, as shown in Fig.~\ref{fig:interval_vs}\footnote{Alternatively, one could also enumerate all the timestamps within the interval and use different $\mathbf{OP_t}$ to project the subject to multiple sets of entities, each of which is specific for one timestamp. Subsequently, an intersection operator again is performed over all the sets of entities obtained from $\mathbf{OP_r}$ and $\mathbf{OP_t}$ in the previous step. However, in spite of its efficiency, this practice is hard to implement in mini-batch training manner since time intervals in different statements usually have varying duration and thus contain different number of timestamps.}. 


\subsection{Implementation in a Vector Space}
In order to implement this idea in a vector space, two key points are 1) how to model a set of answers returned by a KBC model and 2) how to instantiate two projectors and one intersection operator. 

Prior KBC models are incapable of directly representing a set of answer entities in a vector space. Instead, they usually represent entities and relations as single points in the vector space and model point-to-point projections, e.g., TransE. 
Inspired by QUERY2BOX~\cite{ren2019query2box}, which is used to deal with complex queries that involve conjunctions, existential quantifiers, and disjunctions, we introduce the idea of boxes in the vector space and thus name the proposed framework TIME2BOX. The reasons for adopting boxes are three-fold. First, boxes are containers that can naturally model a set of answer entities they enclose. Second, finding the intersection set among sets of entities amounts to finding the intersected area over boxes similar to the concept of Venn diagram. Third, the result of performing an intersection operation over boxes is still a box, which makes it possible to deal with statements of different types in a unified framework. 

In TIME2BOX, each entity $e \in E$, relation $r \in R$, and timestamp $t \in T$ ($T$ is the set of all discrete timestamps in a TKB) are initialized as vector embeddings $\mathbf{e} \in \mathbb{R}^d$, $\mathbf{r} \in \mathbb{R}^d$, and $\mathbf{t} \in \mathbb{R}^d$. $S_r$, $S_t$, and $S_{inter}$ refer to sets of entities and thus are modeled by boxes, represented as box embeddings in the vector space. In the following, we first introduce the definition of box embeddings and then introduce main components of modeling and reasoning.

\subsubsection{Box Construction and Reasoning}

\paragraph{Box embeddings:} 
Mathematically, they are axis-aligned hyper-rectangle in a vector space, which can be determined by the position of the box (i.e., a center point) and its length (i.e., offsets).  Formally, in a vector space $\mathbb R^d$, a box can be represented by $\mathbf{b}$=(Cen($\mathbf{b}$), Off($\mathbf{b}$)), where $\text{Cen}(\mathbf{b}) \in \mathbb{R}^d$ is its center point and $\text{Off}(\mathbf{b}) \in \mathbb{R}_{\ge 0}^d$ specifies the length/2 of the box in each dimension. If an entity belongs to a set, its entity embedding is modeled as a point inside the box of the set. The interior of a box in the vector space can be specified by points inside it:
\begin{equation}
    box_\mathbf{b} = \{\mathbf{e} \in \mathbb R^d: \text{Cen}(\mathbf{b})-\text{Off}(\mathbf{b}) \preceq \mathbf{e} \preceq \text{Cen}(\mathbf{b})+\text{Off}(\mathbf{b}))\}
\end{equation}
where $\preceq$ denotes element-wise inequality.

\paragraph{Projection operators in a vector space}
In previous work, relations are commonly assumed to be projectors that transform a subject embedding to an object embedding in terms of \textit{points} in a vector space, e.g., TransE~\citep{bordes2013translating} and RotatE~\citep{sun2019rotate}. Here we adopt a similar idea but take both relations and timestamps as projectors ($\mathbf{OP_r}$ and $\mathbf{OP_t}$) to project a subject to a set of entities in $S_r$ -- represented as a time-agnostic \textit{box} $\mathbf{b}_{S_r}$ and to a set of entities in $S_t$ -- represented as a time-aware \textit{box} $\mathbf{b}_{S_t}$, respectively, which are illustrated in Fig.~\ref{fig:time_reasoning_process}.  

The center of a box can be defined as the resulting embedding after applying a projection operator ($\mathbf{OP_r}$ or $\mathbf{OP_t}$) on the subject embedding. The centers of $\mathbf{b}_{S_r}$ and $\mathbf{b}_{S_t}$ can be formulated as below:
\begin{align}
Cen(\mathbf{b}_{S_r}) = \mathbf{e} \odot \mathbf{r}; ~~
Cen(\mathbf{b}_{S_t}) = \mathbf{e} \otimes \mathbf{t}
\label{equ:center-proj}
\end{align}
where $\odot \mathbf{r}$ and $\otimes \mathbf{t}$ are projectors $\mathbf{OP_r}$ and $\mathbf{OP_t}$, respectively. Theoretically, projection operators could be instantiated by any projector in existing KBC models, such as element-wise addition in TransE~\citep{bordes2013translating}
, element-wise product in DistMult~\citep{yang2014embedding}
, and Hadamard product in RotatE~\citep{sun2019rotate}.
Even though $\mathbf{OP_r}$ and $\mathbf{OP_t}$ can be different, we choose the same projector for both and implement two TIME2BOX models by taking element-wise addition and element-wise product as operators by following TransE and DistMult, respectively. Accordingly, these two models are named as TIME2BOX-TE and TIME2BOX-DM.

Ideally, the size of the box $\mathbf{b}_{S_r}$ should be determined by both the subject entity and the relation, since the box contains all object entities that satisfy a query in the form of (\textit{s, r, ?o}). The same applies to $\mathbf{b}_{S_r}$. However, as the entity space is usually large in a KB, introducing entity-specific parameters would result in high computational cost. Therefore, in practice, $\text{Off}(\mathbf{b}_{S_r})$ and $\text{Off}(\mathbf{b}_{S_t})$ are only determined by the relation $r \in R$ and the timestamp $t \in T$, respectively. Put differently, the size of $\mathbf{b}_{S_r}$ and $\mathbf{b}_{S_t}$ are initialized based on $r$ and $t$, which are learned through training.

\paragraph{Intersection Operators in a vector space}
An intersection operator aims to find the intersection box $\mathbf{b}_{inter}=\text{(Cen}(\mathbf{b}_{inter}), \text{Off} (\mathbf{b}_{inter}))$ of a set of box embeddings $\mathbf{B} = \{\mathbf{b}_{S_r}, \mathbf{b}_{S_t1}, ..., \mathbf{b}_{S_tn}\}$ obtained from the previous step. The intersection operator should be able to deal with $\mathbf{B}$ of different sizes, as required in Fig.~\ref{fig:time_reasoning_process}. Thus, both $\text{Cen}(\mathbf{b}_{inter})$ and $\text{Off} (\mathbf{b}_{inter})$ are implemented by using attention mechanisms. \sloppy
Following the idea in ~\citet{bahdanau2014neural}, the center point $\text{Cen(}\mathbf{b}_{inter}\text{)}$ is calculated by performing element-wise attention over the centers of boxes in $\mathbf{B}$. 
This can be formulated as follows:
\begin{equation}
    \text{Cen}(\textbf{b}_{inter}) = \sum_i \text{softmax}(\text{NN}(\text{Cen}(\mathbf{b}_{i})) \odot \text{Cen}(\mathbf{b}_{i})
    \label{eq:cent_inter}
\end{equation}
where NN is a one-layer neural network  and $\mathbf{b}_i \in \mathbf{B}$. 

Since the intersection box $\mathbf{b}_{inter}$ must be smaller than any of the box in $\mathbf{B}$, we use element-wise min-pooling to make sure the new box must be shrunk and perform DeepSets~\citep{zaheer2017deep} over all the  Off($\mathbf{b}_i$) ($\mathbf{b}_i \in \mathbf{B}$) to downscale $\mathbf{b}_{inter}$ \citep{ren2019query2box}. This can be written as below:
\begin{equation}
\text{Off(}\mathbf{b}_{inter}\text{)} = \text{Min}(\textbf{Off}) \odot \sigma(\text{DeepSets}(\textbf{Off}))
\end{equation}
where $\text{DeepSets(} \{\mathbf{x_1}, \mathbf{x_2}, ..., \mathbf{x_n}\}\text{)}=\text{MLP}(1/n)\cdot \sum_i^n\text{MLP(} \mathbf{x_i}\text{)}$, $\sigma$ denotes the sigmoid function, and $\textbf{Off}=\{\text{Off}(\mathbf{b}_i):\mathbf{b}_i \in \mathbf{B}\}$.
\subsection{Optimization Objective}
For a query, TIME2BOX aims to pull correct entity embedding into the final box $\mathbf{b}_{inter}$ while pushing incorrect entity embedding far away from it.  The distance-based loss proposed by ~\citet{sun2019rotate} satisfies this need :
\begin{equation}
    Loss = -log \; \sigma(\gamma - D(\mathbf{o}, \mathbf{b}_{inter})) - \frac{1}{k}\sum_{i=1}^{k}log \; \sigma(D(\mathbf{o}^{\prime}, \mathbf{b}_{inter})- \gamma)
    \label{eq:loss}
\end{equation}
where $\sigma$ is the sigmoid function, $\gamma$ is a fixed margin, $\mathbf{o}$ is the embedding of a positive entity to the given query, and $k$ is the number of negative samples $\mathbf{o^{\prime}}$. $ D(\mathbf{o}, \mathbf{b}_{inter})$ measures the distance between entity $\mathbf{o}$ and the final box $\mathbf{b_{inter}}$. With the size of a box being considered, the distance is divided into two parts: outside distance $D_{outside}(\mathbf{o}, \mathbf{b}_{inter})$ and inside distance $D_{inside}(\mathbf{o}, \mathbf{b}_{inter})$. For cases when $\mathbf{o}$ is outside of $\mathbf{b}_{inter}$, the former refers to the distance of an entity embedding $\mathbf{o}$ to the boundary of the box $\mathbf{b}_{inter}$, and the latter calculates the distance between the box's center $Cen(\mathbf{b}_{inter})$ and its boundary.
This can be formalized as below:
\begin{equation}
    D(\mathbf{o}, \mathbf{b}_{inter}) = \alpha \cdot D_{inside}(\mathbf{o}, \mathbf{b}_{inter}) + D_{outside}(\mathbf{o}, \mathbf{b}_{inter})
\end{equation}
where $\alpha \in [0, 1]$. When $\alpha = 0$, it means that a positive entity is required to be in a $\mathbf{b}_{inter}$, but its distance to the center is not as important. $D_{inside}(\mathbf{o}, \mathbf{b}_{inter})$ and $D_{outside}(\mathbf{o}, \mathbf{b}_{inter})$ are written as:

\begin{align*}
     D_{inside}(\mathbf{o}, \mathbf{b}_{inter}) &= \|\text{Cen}(\mathbf{b}_{inter })-\text{Min}(\mathbf{b}_{max}, \text{Max}(\mathbf{b}_{min}, \mathbf{o}))\|_1 \\
     D_{outside}(\mathbf{o}, \mathbf{b}_{inter}) &= \| \text{Max}(\mathbf{o}-\mathbf{b}_{max}, \mathbf{0})+ \text{Max}(\mathbf{b}_{min}-\mathbf{o}, \mathbf{0})\|_1
\end{align*}
where $\mathbf{b}_{min}=\text{Cen}(\mathbf{b}_{inter})-\text{Off}(\mathbf{b}_{inter})$ and $\mathbf{b}_{max}=\text{Cen}(\mathbf{b}_{inter})+\text{Off}(\mathbf{b}_{inter})$ are embeddings of the bottom left corner and the top right corner of $\mathbf{b}_{inter}$, respectively.

Compared to answering atemporal queries, finding correct answers to temporal ones is more challenging. Therefore, the loss function should reward more in the optimization direction that is capable of correctly answering temporal queries. For a given query $q_i$, we use $\frac{1}{n_{q_i}}$, where $n_{q_i}$ is the number of correct answers to $q_i$ that appear in training as a weight to adjust the loss. The core idea here is that time-aware queries often are satisfied with fewer answers, and, thus, are harder to answer compared to atemporal queries.

\subsection{Time Negative Sampling} \label{secsec:time_negative}
Entity negative sampling is widely used in KBC. For a positive sample $(s, r, o)$, negative samples are constructed by replacing $o$ with other entities $o^{\prime}$, ensuring that $(s, r, o^{\prime})$ must not appear in training set. In this paper, we  adopt this strategy so that the model is able to learn the association between entities, relations, and time occurring in a positive sample by distinguishing the correct answers from the negative samples. Moreover, for time-aware statements, we perform time negative sampling, which corrupts a statement $(s, r, o, t)$ by replacing $t$ with a number of timestamps~$t^{\prime}$. This is important for statements where only start time or end time is available. As shown in Fig.~\ref{fig:time_reasoning_process}, the proposed architecture cannot distinguish those statements from time instant-based statements. But time negative sampling can mitigate this issue to some degree. The following is used for time negative sampling concerning different types of statements ($st$ and $et$ are short for start and end time):
\begin{equation}
    T^{\prime} = 
    \begin{cases}
    \{t^{\prime}\in T: (s, r, o, t^{\prime}) \notin \mathcal{G}\} & (s, r, o, t) \\
    \{t^{\prime}\in T: (s, r, o, t^{\prime}) \notin \mathcal{G}, t^{\prime} <st\} & (s, r, o, I_{st}^{-}) \\
    \{t^{\prime}\in T: (s, r, o, t^{\prime}) \notin \mathcal{G}, t^{\prime} >et\} & (s, r, o,I_{-}^{et}) \\
    \{t^{\prime}\in T: (s, r, o,t^{\prime}) \notin \mathcal{G},  t^{\prime} \notin T^{et}_{st}\} & (s, r, o, I_{st}^{et}) \\
    \end{cases}
\end{equation}
where $T^{et}_{st}$ denotes a set of time points within the interval $I_{st}^{et}$.

\subsection{Time Smoothness Regularizer}
Time is continuous. We may expect that neighboring timestamps would have similar representations in the vector space. Following~\citet{lacroix2019tensor}, we penalize time difference between embeddings of two consecutive timestamps by using $L_2$:
\begin{equation}
    \Lambda(T) = \frac{1}{|T|-1}\sum_{i=1}^{|T|-1}\|\mathbf{t}_{i+1}-\mathbf{t}_i\|_2^2
\end{equation}
During the training step, for batches with temporal statements, we add this regularizer with a weight scalar $\beta$ to the loss function in Eq.~\ref{eq:loss}, where $\beta$ specifies the degree of penalization.

\section{Evaluation metrics IN Time prediction} \label{sec:eval_metrics}

\paragraph{Time Interval Evaluation}
$gIOU$~\cite{rezatofighi2019generalized} and $aeIOU$~\citep{jain-etal-2020-temporal} are two evaluation metrics recently adopted in time interval prediction. Both are built on \textit{Intersection Over Union} that is commonly used for bounding box evaluation in Computer Vision. 

\begin{figure}[]
    \centering
    \includegraphics[width=0.5\textwidth]{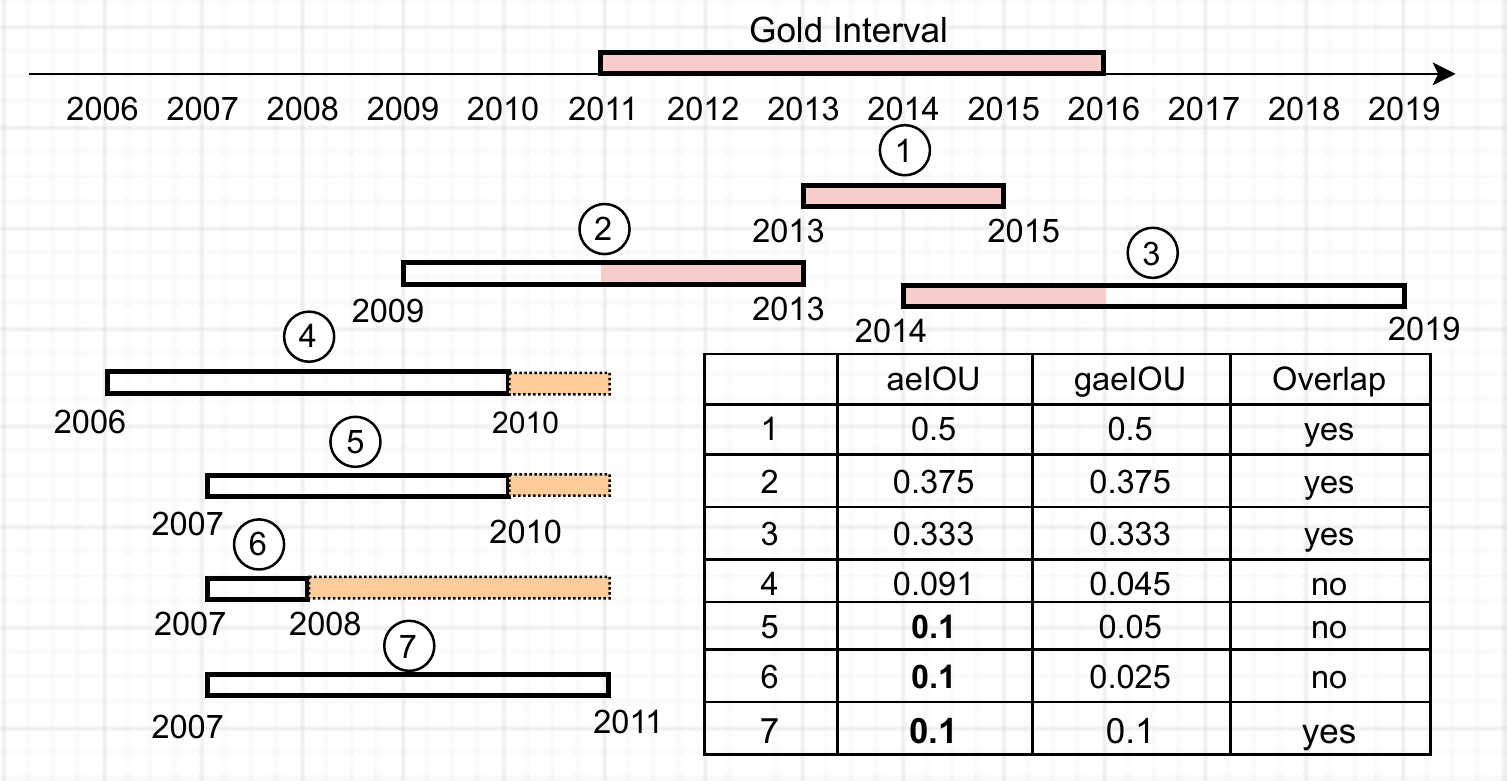}
    \caption{Evaluation Comparison between aeIOU and gaeIOU on different predicted intervals. Suppose a gold interval is [2011, 2016], seven possible predicted intervals are represented as rectangles in black. Intersections between the predicted and the gold are in pink and gaps are in orange if no overlap exists. Notably, gaeIOU is able to distinguish these predictions while aeIOU fails to do so. }
    \label{fig:interval_eval}
    \vspace{-0.5cm}
\end{figure}

The idea of $gIOU$ is to compare the intersection between a predicted interval and a gold interval against the maximal extent that the two intervals may expand. It can be formulated as below:
\begin{equation}
\begin{split}
  gIOU(I^{gold}, I^{pred}) = 
    \frac{D(I^{gold} \bigcap I^{pred})}{D(I^{gold}  \bigcup I^{pred})} - \\ \frac{D(I^{gold} \biguplus I^{pred}\setminus (I^{gold}  \bigcup I^{pred}))}{D(I^{gold}  \biguplus I^{pred})} \in (-1, 1]
\end{split}
\end{equation}
\sloppy
where $I^{gold} \bigcap I^{pred}$ is the overlapping part of two intervals, $I^{gold} \biguplus I^{pred}$ denotes the shortest contiguous interval (hull) that contains both $I^{gold}$ and $I^{pred}$. As shown in Fig.~\ref{fig:interval_eval}, if $I^{gold}=[2011, 2016]$ and $I^{pred}=[2009, 2013]$, then $I^{gold} \bigcap I^{pred}=[2011, 2013]$ and $I^{gold} \biguplus I^{pred}=[2009, 2016]$. $D(I) = I_{max}-I_{min}+1$ is the number of time points at a certain granularity (e.g., year in this paper) during the time interval $I$. 

Compared to $gIOU$, affinity enhanced IOU, denoted as $aeIOU$, provides a better evaluation in case of non-overlapping intervals and outputs scores in $[0, 1]$.
It can be written as follow:
\begin{equation}
    aeIOU(I^{gold}, I^{pred}) = 
    \begin{cases}
    \frac{D(I^{gold} \bigcap I^{pred})}{D(I^{gold}  \biguplus I^{pred})} & D(I^{gold} \bigcap I^{pred}) > 0 \\
    \frac{1}{D(I^{gold}  \biguplus I^{pred})} & otherwise
    \end{cases}
    \label{eq:aeIOU}
\end{equation}
However, we notice that $aeIOU$ cannot tell some cases apart. As illustrated in Fig.~\ref{fig:interval_eval}, $aeIOU$ results in the same scores for \textcircled{5}, \textcircled{6}, and \textcircled{7} when compared to the gold interval$[2011,2016]$. Intuitively one would assume that \textcircled{7} is better than the others and \textcircled{6} is the least desirable. The former has a one-year intersection between \textcircled{7} and the gold.  For the latter, the gap between \textcircled{5} and the gold is smaller than that between \textcircled{6} and the gold, despite the fact that neither \textcircled{5} and \textcircled{6} overlaps with the gold. Its failure lies in that it does not consider the gap between the gold and the predicted interval in case of no overlap.

In the following, we take both the hull and the intersection/gap between a gold interval and a predicted interval into the design of the metric. The intuition is that when the size of the hull remains the same, the metric score of a predicted interval \textit{decreases} with a larger gap to the gold in case of no overlap and \textit{increases} with a larger intersection. $aeIOU$ is therefore generalized to $gaeIOU$ as below:
\begin{equation}
    gaeIOU(I^{gold}, I^{pred}) = 
    \begin{cases}
    \frac{D(I^{gold} \bigcap I^{pred})}{D(I^{gold}  \biguplus I^{pred})} & D(I^{gold} \bigcap I^{pred}) > 0 \\ \\
    \frac{D^{\prime}(I^{gold}, ~I^{pred})^{-1}}{D(I^{gold}  \biguplus I^{pred})} & otherwise
    \end{cases}
    \label{eq:aeIOU}
\end{equation}
where $D^{\prime}(I^{gold}, ~I^{pred}) = max(I^{gold}_{min},I^{pred}_{min})-min(I^{gold}_{max},I^{pred}_{max})+1$ is the length of the gap. 

 Accordingly, the Property ($P$) that a good evaluation metric must satisfy can be rewritten as: if predicted intervals (partially) overlap with the gold interval with the same size, then the prediction having a smaller hull with the gold interval should be awarded more by $M$; if there is no overlap, the prediction that has a smaller hull and a narrower gap with the gold should be scored higher by $M$. It can be formalized as below:

\sloppy
\textbf{Property P}: In case of $D(I^{gold}  \bigcap I^{pred1}) = D(I^{gold} \bigcap I^{pred2}) \neq 0$, $M(I^{gold}, I^{pred1})>M(I^{gold}, I^{pred2})$ if and only if $D(I^{gold}\bigcup I^{pred1})<D(I^{gold} \bigcup I^{pred2})$. 
\\
\sloppy
In case of non-overlapping, $M(I^{gold}, I^{pred1}) > M(I^{gold}, I^{pred2})$ if and only if $D(I^{gold}  \bigcup I^{pred1}) \cdot D^{\prime}(I^{gold}  \bigcap I^{pred1}) > D(I^{gold}  \bigcup I^{pred2}) \cdot D^{\prime}(I^{gold}  \bigcap I^{pred2}))$.

It follows that $gaeIOU$ satisfies Property P, whereas $aeIOU$ does not satisfy it; see  Fig.~\ref{fig:interval_eval}.

\section{Experiment} \label{sec:experiment}
Our goal here is to evaluate TIME2BOX in both link prediction and time prediction tasks. For a test sample $(s, r, ?o, t^{*})$, we replace $?o$ with each entity $o^\prime \in E$ and use $log\sigma(\gamma-D(\mathbf{o^\prime}, \mathbf{b}_{inter}))$, a variation of the inverse distance used in Eq.~\ref{eq:loss}, as scores for link prediction. Entities that have higher scores are more likely to form new links. Likewise, in terms of time prediction, for a query $(s, r, o, ?I)$, we first replace $I$ with each timestamp $t \in T$ and calculate its score. Then we use the greedily coalescing method proposed in~\citet{jain-etal-2020-temporal} to generate time intervals as predictions. 

\subsection{Datasets}
We report experimental results using two TKBC datasets, which both are rooted in Wikidata. WIKIDATA12k is a widely used benchmark dataset in TKBC where each statement is associated with a time ``interval''~\citep{dasgupta2018hyte}. Such an ``interval'', in fact, could be a time instant, where start time and end time are the same, a left/right-open interval, or a closed interval. Note that this dataset excludes statements that do not have known temporal scopes in Wikidata, although they may be time-dependent and useful in TKBC, as discussed in Section~\ref{sec:intro}. The other dataset is a subset of WIKIDATA432k proposed by ~\citet{lacroix2019tensor}, which is the only TKB dataset where the start time, end time, or both of a statement can remain unspecified. Although this dataset is more appropriate for a TKBC problem, there are two limitations. First, it poses a computational burden as it contains 432k entities and 407 relations, consisting of 7M tuples in the training set. Second, there are several mistakes in the temporal information. For instance, 2014 was written as 2401. We extract a subgraph, named as  WIKIDATA114k, and correct temporal information by checking it against Wikidata. More details about data pre-processing and statistics are in Appendix A (All Appendices are available online\footnote{\href{https://github.com/ling-cai/Temporal-KG/blob/28db101fa09853f761547f99d57f0400189bc413/Paper/Time2Box_Appendix.pdf}{Link to online Appendices.}}). Since our focus is on generic knowledge bases, we do not consider event-based datasets, such as ICEWS14 and ICEWS05-15, in which each statement is associated with a timestamp. 

\subsection{Baselines and Model Variants}
In the following experiments, we regard TIME2BOX-TE as our main model, in which both the relation operator and the time projector are instantiated as an element-wise addition. It is denoted as TIME2BOX in resulting tables. We compare it against two SOTAs in TKBC: TNT-Complex and TIMEPLEX base model by using the implementation in~\citet{jain-etal-2020-temporal}, both of which are based on the time-agnostic KBC model: ComplEx~\citep{trouillon2016complex}.


In addition to comparison with existing SOTAs, we also conduct an ablation study, in which several variants of the proposed model are compared: (1) TIME2BOX-SI, short for Sample Interval: for a closed interval-based statement, this variant randomly samples a sub time interval from a given interval at each training step and train it as shown in Fig.~\ref{fig:interval_vs}. (2) TIME2BOX-TR: previous works in TKBC often explicitly fused relations with time information to obtain time-aware relations and empirically demonstrated its effectiveness~\citep{jain-etal-2020-temporal,lacroix2019tensor, garcia-duran-etal-2018-learning}. We also explicitly model the association between relations and time as a new point $p_{rt}=\mathbf{r}+\mathbf{t}$ in the vector space and incorporate it into Eq.~\ref{eq:cent_inter} to help locate the intersection box. 
(3)TIME2BOX-DM: this variant implements the relation and time projectors as an element-wise product in real space as DistMult does. 
(4) TIME2BOX-TNS: this variant is used to test the effect of time negative samples, in which we replace a number of entity negative samples with time negative samples, as introduced in Section~\ref{secsec:time_negative}. 

All these models are trained on statements in training set and evaluated by answering queries where either the object or the time information is missing. Hyper-parameter settings are introduced in Appendix B and comparison of parameters used in different models is summarized in Table 11 in Appendix F. Moreover, we notice there are several limitations in current experimental setups of SOTAs and we detail them in Appendix C.


\subsection{Main Results}
\begin{table}[h]
\resizebox{0.5\textwidth}{!}{\setlength\tabcolsep{1.5pt}\begin{tabular}{lcccc|cccc}
\hline
Datasets      & \multicolumn{4}{c|}{WIKIDATA12k}                                & \multicolumn{4}{c}{WIKIDATA114k}                                                                        \\ \hline
Metrics       & MRR            & MR           & HITS@1         & HITS@10        & MRR                                & MR                               & HITS@1         & HITS@10        \\ \hline
TNT-Complex   & 31.77          & 415          & 19.24          & 51.74          & 49.25                              & 638                              & 41.02          & 66.99          \\
TIMEPLEX base & 34.55          & 302          & 21.91          & 53.25          & 49.99                              & 337                              & 41.25          & 66.10          \\ \hline
TIME2BOX-TR   & 34.99          & 102          & 24.79          & 56.32          & \multicolumn{1}{l}{50.25}          & 85                               & 41.73          & 67.13          \\
TIME2BOX-DM   & 35.90          & 139          & 25.52          & 56.74          & \multicolumn{1}{l}{48.84}          & \multicolumn{1}{l}{284}          & 41.09          & 64.33          \\
TIME2BOX-SI   & 36.79          & \textbf{100} & 27.16          & 56.43          & \multicolumn{1}{l}{50.42}          & \multicolumn{1}{l}{\textbf{139}} & 41.65          & 67.58          \\
TIME2BOX-TNS  & 37.25          & \textbf{100} & \textbf{27.41} & 57.31          & \multicolumn{1}{l}{\textbf{50.55}} & \multicolumn{1}{l}{185}          & \textbf{41.77} & 67.78          \\
TIME2BOX   & \textbf{37.30} & 101          & 27.38          & \textbf{57.36} & 50.49                              & 168                              & 41.69          & \textbf{67.91} \\ \hline
\end{tabular}
\vspace{-1.0cm}
}
\caption{Link prediction evaluation across two datasets.}
\label{tb:eval_link_prediction}
\vspace{-0.8cm}
\end{table}
\paragraph{Link Prediction Task} We report main results of link prediction in Table~\ref{tb:eval_link_prediction}. TIME2BOX and all its variants consistently outperform or are on a par with the performance on SOTAs in terms of MRR, MR, HITS@1 and HITS@10. On WIKIDATA12k, TIME2BOX outperforms TIMEPLEX base by around 3 points in terms of MRR and over 5 points in HITS@1. On WIKIDATA114k, TIME2BOX is 
slightly better than two SOTAs in general for MRR, HITS@1 and HITS@10. In addition, we notice that TIME2BOX beats SOTAs by large margins in time interval-based link prediction, as shown in Table 8 in Appendix D. Our method improves around 20 and 7 HITS@1 points in terms of half-open interval-based link prediction and closed interval-based link prediction on WIKIDATA12k, respectively. On WIKIDATA114k TIME2BOX improves around 6 and 4 HITS@1 points, respectively. 

Another critical observation in Table~\ref{tb:eval_link_prediction} is the substantial improvements of using TIME2BOX in terms of MR on both datasets. TIME2BOX returns an MR of 100 and 139 on WIKIDATA12k and WIKIDATA114k, respectively and TIMEPLEX base obtains 302 and 337 for MR on both datasets. It indicates that TIME2BOX is capable of giving a fair rank for a gold answer to any test query on average. This is likely because of the idea of using boxes to constraint the potential answer set. As a time-agnostic box is optimized towards embracing entities satisfying atemporal queries of the form ($s, r, ?o$) in the learning process, boxes implicitly manage to learn common characteristics of the satisfied $?o$. Therefore, TIME2BOX is less likely to output extremely bad predictions.  Examples in Section~\ref{subsec:quality_study} exemplify this hypothesis. 

\paragraph{Time Prediction Task} Table~\ref{tb:eval_time_prediction_12k} and Table~\ref{tb:eval_time_prediction_114k} summarizes the results for two datasets. On both datasets, TIME2BOX and its variants consistently outperform SOTAs by significant margins. Specifically, TIME2BOX improves over TIMEPLE by about 5.56, 7.25, and 4.87 points with respect to gIOU@1, aeIOU@1, and gaeIOU@1, respectively, on WIKIDATA12k. As for WIKIDATA114k, despite subtle improvements in link prediction, the advancement of TIME2BOX is more pronounced in time prediction, which shows that it gains 8.7, 5.87, and 4.66 points on gIOU@1, aeIOU@1, and gaeIOU@1, respectively. Furthermore, the improvements on gaeIOU@10 are much more notable with gains of 15.81 and 11.07 points on the two datasets, respectively. 

\begin{table}[h]

\resizebox{0.5\textwidth}{!}{
\setlength\tabcolsep{1.5pt}
\begin{tabular}{|lcc|cc|cc|}
\hline
Datasets     & \multicolumn{6}{c|}{WIKIDATA12k}                                                                    \\ \hline
Metrics      & gIOU@1         & gIOU@10        & aeIOU@1        & aeIOU@10       & gaeIOU@1       & gaeIOU@10      \\ \hline
TNT-Complex  & 31.44          & 55.18          & 18.86          & 40.94          & 11..01         & 29.51          \\
TIMEPLEX base    & 35.63          & 60.86          & 18.60          & 37.75          & 12.61          & 32.63          \\ \hline
TIME2BOX-TR  & 39.63          & 67.83          & 23.47          & 44.64          & 15.87          & 41.53          \\
TIME2BOX-DM  & 38.78          & 62.44          & 21.91          & 41.55          & 14.94          & 37.14          \\
TIME2BOX-SI  & 39.68          & 65.30          & 23.66          & 42.16          & 16.09          & 38.54          \\
TIME2BOX-TNS & \textbf{42.30} & \textbf{70.16} & \textbf{25.78} & \textbf{50.04} & \textbf{17.41} & \textbf{47.54} \\
TIME2BOX     & 41.20          & 68.53          & 24.70          & 46.05          & 16.98          & 43.08          \\ \hline
\end{tabular}
}
\caption{Time prediction evaluation on WIKIDATA12k.}
\label{tb:eval_time_prediction_12k}
\vspace{-1.0cm}
\end{table}

\begin{table}[h]
\resizebox{.5\textwidth}{!}{
\setlength\tabcolsep{1.5pt}
\begin{tabular}{|lcc|cc|cc|}
\hline
Datasets     & \multicolumn{6}{c|}{WIKIDATA114k}                                                                                                                                                       \\ \hline
Metrics      & \multicolumn{1}{l}{gIOU@1} & \multicolumn{1}{l|}{gIOU@10} & \multicolumn{1}{l}{aeIOU@1} & \multicolumn{1}{l|}{aeIOU@10} & \multicolumn{1}{l}{gaeIOU@1} & \multicolumn{1}{l|}{gaeIOU@10} \\ \hline
TNT-Complex  & 27.94                      & 48.31                        & 16.18                       & 35.32                         & 7.31                         & 23.68                          \\
TIMEPLEX base    & 29.31                      & 57.68                        & 18.56                       & 36.70                         & 12.53                        & 32.47                          \\ \hline
TIME2BOX-TR  & 37.49                      & 67.95                        & 25.05                       & 49.02                         & 15.41                        & 45.72                          \\
TIME2BOX-DM  & 35.88                      & 66.62                        & 24.33                       & 48.03                         & 14.89                        & 44.48                          \\
TIME2BOX-SI  & 34.02                      & 62.89                        & 23.10                       & 44.74                         & 14.07                        & 40.05                          \\
TIME2BOX-TNS & 37.31                      & 66.91                        & 25.07                       & 48.18                         & 15.57                        & 44.66                          \\
TIME2BOX     & \textbf{38.01}             & \textbf{71.29}               & \textbf{24.42}              & \textbf{50.07}                & \textbf{15.88}               & \textbf{47.77}                 \\ \hline
\end{tabular}
}
\caption{Time prediction evaluation on WIKIDATA114k.}
\label{tb:eval_time_prediction_114k}
\vspace{-0.8cm}
\end{table}

\subsection{Qualitative Study} \label{subsec:quality_study}
Table~\ref{tb:timestamped_query_example} showcases examples of timestamp-based link prediction on WIKIDATA12k. The comparison between TIMEPLEX base and TIME2BOX reveals that TIME2BOX is able to learn common characteristics of entities by adopting boxes. For instance, the predicted top 10 returned by TIME2BOX are possible affiliations (e.g., institutes, colleges, universities) in the first query and are countries in the second query. By contrast, TIMEPLEX base returns a mixture of entities with distinct classes for both queries. Furthermore, Table~\ref{tb:timeinterval_query_example} shows an example of time interval-based link prediction, in which TIME2BOX is able to consistently output correct predictions across time and precisely discern the changes of objects over time (i.e., the correct answer shifts from Russian Empire to Ukrainian People's Republic in 1916), while TIMEPLEX base fails. This can be attributed to the ability of TIME2BOX to capture the order of timestamps and the idea of temporal boxes as a constraint over potential answer entities. Hence, answer entities that are true in two consecutive years can be enclosed in the intersection of temporal boxes. 

\begin{table}[]
\resizebox{0.55\textwidth}{!}{\setlength\tabcolsep{1.5pt}\begin{tabular}{|c|l|}
\hline
\multicolumn{2}{|c|}{\textit{\textbf{Query Example 1: (Yury Vasilyevich Malyshev, educatedAt, ?o, 1977)}}} \\ \hline
TIMEPLEX base & \multicolumn{1}{c|}{TIME2BOX} \\ \hline
\multicolumn{1}{|l|}{\begin{tabular}[c]{@{}l@{}}1. Bauman Moscow State Technical University,\\ 2. Gold Star,\\ 3. Communist Party of the Soviet Union,\\ 4. Order of Lenin,\\ 5. S.P. Korolev Rocket and Space Corporation Energia,\\ 6. Hero of the Soviet Union,\\ 7. \underline{\textbf{Gagarin Air Force Academy,}}\\ 8. Balashov Higher Military Aviation School of Pilots,\\ 9. Ashok Chakra,\\ 10. Heidelberg University\end{tabular}} & \begin{tabular}[c]{@{}l@{}}1. Bauman Moscow State Technical University,\\ \underline{\textbf{2. Gagarin Air Force Academy,}}\\ 3. S.P. Korolev Rocket and Space Corporation Energia,\\ 4. Saint Petersburg State Polytechnical University,\\ 5. University of Oxford,\\ 6. Saint Petersburg State University,\\ 7. Steklov Institute of Mathematics,\\ 8. Leipzig University,\\ 9. Heidelberg University,\\ 10. Moscow Conservatory\end{tabular} \\ \hline
\multicolumn{2}{|c|}{\textit{\textbf{Query Example 2: (Pedro Pablo Kuczynski, countryOfCitizenship,?o, 2015)}}}\\ \hline
TIMEPLEX base & \multicolumn{1}{c|}{TIME2BOX} \\ \hline
\multicolumn{1}{|l|}{\begin{tabular}[c]{@{}l@{}}1. doctor honoris causa,\\ 2. President of Peru,\\ 3. Minister of Economy and Finance of Peru,\\ 4. Grand Cross of the Order of the Sun of Peru,\\ 5. President of the Council of Ministers of Peru,\\ 6. World Bank,\\ 7. Serbia,\\ 8. Royal Spanish Academy,\\ 9. Meurthe-et-Moselle,\\ 10. Norwegian Sportsperson of the Year\end{tabular}}                           & \begin{tabular}[c]{@{}l@{}}1. France,\\ 2. Germany,\\ \underline{\textbf{3. United States of America,}}\\ 4. Austria,\\ 5. Romania,\\ 6. United Kingdom,\\ 7. Poland,\\ 8. Kingdom of Italy,\\ 9. Russian Soviet Federative Socialist Republic,\\ 10. Russian Empire\end{tabular}  \\ \hline
\end{tabular}}
\caption{Examples of timestamp-based link prediction on WIKIDATA12k. Top 10 entities predicted by TIMEPLEX base and TIME2BOX are numbered, where 1 denotes Top One. Correct answers are in bold.}
\label{tb:timestamped_query_example}
\vspace{-0.8cm}
\end{table}

\begin{table}[]
\resizebox{0.5\textwidth}{!}{\setlength\tabcolsep{1.5pt}\begin{tabular}{|l|l|l|}
\hline
     & \multicolumn{2}{l|}{Query Example: (Kyiv, country, ?o, {[}1905, 1919{]})}                                                                                                                \\ \hline
     & \multicolumn{2}{l|}{\begin{tabular}[c]{@{}l@{}}Gold Answers: (1){[}1905, 1916{]}-\textgreater Russian Empire; (2){[}1917, 1919{]}-\textgreater{}Ukrainian People's Republic\end{tabular}} \\ \hline
year & Timplex                                                                                   & TIME2BOX                                                                               \\ \hline
1905 & Soviet Union                                                                                    & Russian Empire                                                                         \\ \hline
1906 & Soviet Union                                                                                    & Russian Empire                                                                         \\ \hline
1907 & Russian Empire                                                                                  & Russian Empire                                                                         \\ \hline
1908 & Soviet Union                                                                                    & Russian Empire                                                                         \\ \hline
1909 & Ukrainian Soviet Socialist Republic                                                             & Russian Empire                                                                         \\ \hline
1910 & Soviet Union                                                                                    & Russian Empire                                                                         \\ \hline
1911 & Russian Empire                                                                                  & Russian Empire                                                                         \\ \hline
1912 & Ukrainian Soviet Socialist Republic                                                             & Russian Empire                                                                         \\ \hline
1913 & Soviet Union                                                                                    & Russian Empire                                                                         \\ \hline
1914 & Ukrainian Soviet Socialist Republic                                                             & Russian Empire                                                                         \\ \hline
1915 & Soviet Union                                                                                    & Russian Empire                                                                         \\ \hline
1916 & Ukrainian People's Republic                                                                     & Russian Empire                                                                         \\ \hline
1917 & Ukrainian People's Republic                                                                     & Ukrainian People's Republic                                                            \\ \hline
1918 & Ukrainian People's Republic                                                                     & Ukrainian People's Republic                                                            \\ \hline
1919 & Ukrainian People's Republic                                                                     & Ukrainian People's Republic                                                            \\ \hline
\end{tabular}}
\caption{An example of interval-based link prediction on WIKIDATA12k. For time interval-based link prediction, the current strategy is to discretize intervals to timestamps and average ranks for each timestamp-based prediction result as the final evaluation. Only top 1 predictions are shown here.}
\label{tb:timeinterval_query_example}
\vspace{-1.05 cm}
\end{table}

\subsection{Model Variation Study}
In this section, we report on observations of results about different model variations, which are shown in Table~\ref{tb:eval_link_prediction},~\ref{tb:eval_time_prediction_12k} and ~\ref{tb:eval_time_prediction_114k}. Compared to TIME2BOX-DM, which adopts element-wise product as operators, element-wise addition projectors (TIME2BOX) perform better in link prediction and time prediction on both datasets. Moreover, we observe that explicitly modeling association between time and relation (i.e., TIME2BOX-TR) does not significantly improve the performance of TIME2BOX framework, although it speeds up the convergence at training, indicating that intersection operators are good enough to learn the association between time and relations implicitly. As for different time-aware strategies incorporated in TIME2BOX-SI and TIME2BOX-TNS, we find that on both datasets TIME2BOX-SI does not outperform TIME2BOX, indicating that using one sample strategy (i.e., Fig.~\ref{fig:instant_vs}) is better at modeling time interval-based statements in TIME2BOX. In addition, we find that by incorporating time negative samples, the performance on time prediction can be further improved on WIKIDATA12k, although TIME2BOX-TNS is not superior to TIME2BOX in link prediction. 
\section{Conclusion} \label{sec:conclusion}
In this work, we presented a box-based temporal knowledge graph (TKBC) completion framework (called TIME2BOX) to represent and model statements with different types of validity information (i.e., no time, known start time, known end time, instant, both start and end time) in a vector space. We argued that a TKBC problem can be solved in two steps. First by solving an atemporal KBC problem and then narrowing down the correct answer sets that are only true at the time of interest. Therefore, we  introduced time-agnostic boxes to model sets of answers 
obtained from KBC models. Time-aware boxes are used as a filter to pick out time-dependent answers. 
TIME2BOX outperforms existing TKBC methods in both link prediction and time prediction on two datasets - WIKIDATA12k and WIKIDATA114K. By investigating the model performance on statements with different types of validity information, we found that the improvement of TIME2BOX largely attributes to its better ability to handle statements with interval-based validity information. In the future, we will explore how to incorporate spatial scopes of statements into KGE models, such that KBC can benefit from both spatial and temporal scopes of statements.

\begin{acks}

This work was partially supported by the NSF award 2033521, “KnowWhereGraph: Enriching and Linking Cross-Domain Knowledge Graphs using Spatially-Explicit AI Technologies".
\end{acks}


\bibliographystyle{ACM-Reference-Format}
\bibliography{references}

\clearpage
\begin{appendix}


\section{Data Statistics} \label{ap:data_desp}
\paragraph{Generation of WIKIDATA114k}
 We extracted a sport-centric subgraph from WIKIDADA432k. We first picked out statements where the relation \textit{memberOfSportsTeam} appears and obtained an entity set from those statements. Then we find all the statements that entities obtained from the previous step participate in as our initial subgraph. Finally, we ensure that each entity/relation is associated with at least 5 statements and the time period is restricted to [1883, 2023] for temporal statements, which encloses most of the temporal statements in the initial subgraph. This results in 1.7 million statements with 114k entities and 126 relations, and thus named as WIKIDATA114k. See Table ~\ref{tb:data_stat} for data statistics.
\begin{table}[h]
\begin{tabular}{llll}
\hline
\multicolumn{2}{l}{}      & WIKIDATA12k    & WIKIDATA114k     \\ \hline
\multicolumn{2}{l|}{\#entities}         & 12,544         & 114,351          \\
\multicolumn{2}{l|}{\#relations}        & 24             & 126              \\
\multicolumn{2}{l|}{time period}        & {[}19, 2020{]} & {[}1883, 2023{]} \\ \hline
\multicolumn{1}{l|}{\multirow{6}{*}{train}}  & \multicolumn{1}{l|}{\#all}                & 32,497         & 1,670,969        \\
\multicolumn{1}{l|}{}                        & \multicolumn{1}{l|}{\#time instant}       & 14,099         & 175,637          \\
\multicolumn{1}{l|}{}                        & \multicolumn{1}{l|}{\#start time only}    & 4,089          & 44,809           \\
\multicolumn{1}{l|}{}                        & \multicolumn{1}{l|}{\#end time only}      & 1,273          & 2,164            \\
\multicolumn{1}{l|}{}                        & \multicolumn{1}{l|}{\#full time interval} & 13,035         & 402,135          \\
\multicolumn{1}{l|}{}                        & \multicolumn{1}{l|}{\#no time}            & 0              & 1,046,224        \\ \hline
\multicolumn{1}{l|}{\multirow{6}{*}{valid}}  & \multicolumn{1}{l|}{\#all}                & 4,051          & 11,720           \\
\multicolumn{1}{l|}{}                        & \multicolumn{1}{l|}{\#time instant}       & 1,857          & 1,177            \\
\multicolumn{1}{l|}{}                        & \multicolumn{1}{l|}{\#start time only}    & 322            & 342              \\
\multicolumn{1}{l|}{}                        & \multicolumn{1}{l|}{\#end time only}      & 76             & 11               \\
\multicolumn{1}{l|}{}                        & \multicolumn{1}{l|}{\#full time interval} & 1,796          & 2,655            \\
\multicolumn{1}{l|}{}                        & \multicolumn{1}{l|}{\#no time}            & 0              & 7,535            \\ \hline
\multicolumn{1}{l|}{\multirow{6}{*}{\#test}} & \multicolumn{1}{l|}{\#all}                & 4,043          & 11,854           \\
\multicolumn{1}{l|}{}                        & \multicolumn{1}{l|}{\#time instant}       & 1,844          & 1,219            \\
\multicolumn{1}{l|}{}                        & \multicolumn{1}{l|}{\#start time only}    & 324            & 306              \\
\multicolumn{1}{l|}{}                        & \multicolumn{1}{l|}{\#end time only}      & 56             & 15               \\
\multicolumn{1}{l|}{}                        & \multicolumn{1}{l|}{\#full time interval} & 1,819          & 2,790            \\
\multicolumn{1}{l|}{}                        & \multicolumn{1}{l|}{\#no time}            & 0              & 7,524            \\ \hline
\end{tabular}
\caption{Statistics of these datasets used.}
\label{tb:data_stat}
\end{table}

\section{Hyperparameter Settings}
We tune models by the MRR on the validation set. Grid search is performed over negative samples $k$ = [16, 32, 64, 128], learning rate $lr =$ [0.003, 0.002, 0.001], batch size $b =$ [1500, 2000, 2500, 3000, 3500]; dimension $d =$ [200, 300, 400], and weight for time smoothness regularizer $\beta =$ [0.0, 0.1, 0.001, 0.0001], as shown in Table~\ref{tb:hyperparamters}.\footnote{Experiments are terminated after 10000 steps.} We find that effects of different hyperparameters are minimal except for learning rate as the trained model usually converge to similar MRRs as long as they are trained thoroughly. We also observe that time smoothness regularizer is useful in learning time embeddings on WIKIDATA12k while failing to improve the model on WIKIDATA114k. This may be due to data sparsity with regard to time. As the time span of WIKIDATA114k is much smaller, time information is intensive and thus models are capable of learning temporal order between timestamps implicitly.

\begin{table}[H]
\centering
\begin{tabular}{@{}l|llll|lll@{}}
\toprule
       & \multicolumn{4}{l|}{\#negative samples} & \multicolumn{3}{l}{\# learning rate} \\ \midrule
       & 16       & 32       & 64      & 128     & 0.003      & 0.002      & 0.001      \\ \midrule
MRR    & 36.02    & 36.68    & 37.06   & 37.30   & 36.64      & 36.82      & 37.30      \\
MR     & 97       & 100      & 98      & 101     & 126        & 103        & 101        \\
HITS@1 & 25.96    & 26.81    & 27.25   & 27.38   & 26.83      & 26.73      & 27.38      \\ \midrule
       & \multicolumn{4}{l|}{\#batch size}       & \multicolumn{3}{l}{\# dimension}     \\ \midrule
       & 2000     & 2500     & 3000    & 3500    & 200        & 300        & 400        \\ \midrule
MRR    & 36.71    & 36.87    & 37.30   & 36.78   & 36.12      & 36.88      & 37.30      \\
MR     & 100      & 114      & 101     & 100     & 106        & 101        & 101        \\
HITS@1 & 26.59    & 26.88    & 27.38   & 26.77   & 25.98      & 26.88      & 27.38      \\ \bottomrule
\end{tabular}
\caption{Effects of hyper-parameters on WIKIDATA12k}
\label{tb:hyperparamters}
\end{table}

\section{Experimental Setup} \label{sec:exp_setup}
Upon inspection on implementations of TKBC models, we find there are two common issues. 

First, SOTAs only learn time embeddings for timestamps that appear in training set, which would be problematic at testing. For instance, suppose a sorted (ascending) list of timestamps occurring in training set is [1540, 1569, 1788, 1789, 1790], SOTAs only learn embeddings for these timestamps, while ignoring intermediate timestamps. As a result, they cannot answer queries when the associated time is not in the list, such as (s, r, ?o, \textit{1955}). This problem would be even worse regarding time interval generation. As when we need to grow a time point to a time interval by extending it to the left or the right, we may jump from one year to a year far away from it. For instance, from 1569 to 1540 (left) or 1788 (right). This is not reasonable and thus may severely affect the evaluation on time prediction. In order to address this issue, we enumerate all the time points in the time span of the training set with a fixed granularity (i.e., year) and use them for all models at training periods. 

The other issue is about the evaluation of link prediction task on time interval-based statements (including closed interval-based and left/right-open interval-based statements). In existing works, the evaluation boils down to assessing the correctness of answering a timestamp-based query by randomly picking one timestamp from a set of timestamps within the time interval and then measuring the performance on the newly generated query (i.e., the timestamp-based query). However, this is problematic. For closed interval-based samples, the evaluation results may vary from randomly sampled timestamps and thus may not be stable. For left/right-open interval-based statements, it is more severe. For instance, for a left-open interval-based test sample \textit{(Albert Einstein, educatedAt, ?o, [-, 1905])}, \citet{lacroix2019tensor} randomly pick a year before 1905, say 1000, and evaluate whether a  model can output the correct answer (\textit{University of Zurich}) to the new query \textit{(Albert Einstein, educatedAt, ?o, 1000)}. Clearly, there is no correct answer at all since he was born in 1879. Therefore, the evaluation on such test samples may not be plausible. In order to address these issues,
for a closed interval-based sample, we enumerate all the time points in the interval and do evaluation on each time point separately. Then we use the average performance over them as the overall evaluation. For the latter, we only consider the known endpoint in an interval, namely $(s, r, ?o, st)$ for right-open cases and  $(s, r, ?o, et)$ for left-open cases.

\section{Link Prediction Performance by types of validity information} \label{ap:link_prediction_by_types}
Table~\ref{tb:eval_time_prediction_bytype} shows the comparison between different methods in terms of different types of validity information.
\begin{table}[!htbp]
\resizebox{0.5\textwidth}{!}{
\setlength\tabcolsep{1.5pt}
\begin{tabular}{lllllllll}
\hline
Datasets                     & \multicolumn{8}{c}{WIKIDATA12k}                \\ \hline
Types                        & \multicolumn{2}{c|}{Time Interval (O)}               & \multicolumn{2}{c|}{Time Interval (C)}              & \multicolumn{2}{c|}{Time Instant}                   & \multicolumn{2}{c}{No Time}                   \\ \hline
Methods                      & TIMEPLEX base  & \multicolumn{1}{l|}{TIME2BOX}       & TIMEPLEX base & \multicolumn{1}{l|}{TIME2BOX}       & TIMEPLEX base & \multicolumn{1}{l|}{TIME2BOX}       & TIMEPLEX base        & TIME2BOX              \\ \hline
MRR                          & 46.74          & \multicolumn{1}{l|}{\textbf{51.48}} & 25.30         & \multicolumn{1}{l|}{\textbf{28.44}} & 41.11         & \multicolumn{1}{l|}{\textbf{43.13}} & \multicolumn{1}{c}{-} & \multicolumn{1}{c}{-} \\
MR                           & 203            & \multicolumn{1}{l|}{\textbf{68}}    & 273           & \multicolumn{1}{l|}{\textbf{84}}    & 350           & \multicolumn{1}{l|}{\textbf{125}}   & \multicolumn{1}{c}{-} & \multicolumn{1}{c}{-} \\
HITS@1                       & 19.21          & \multicolumn{1}{l|}{\textbf{41.05}} & 11.54         & \multicolumn{1}{l|}{\textbf{18.5}}  & 32.6          & \multicolumn{1}{l|}{\textbf{33.30}} & \multicolumn{1}{c}{-} & \multicolumn{1}{c}{-} \\ \hline
\multicolumn{1}{c}{Datasets} & \multicolumn{8}{c}{WIKIDATA114k}                                                                                                                                                                                 \\ \hline
\multicolumn{1}{c}{Types}    & \multicolumn{2}{c|}{Time Interval (O)}               & \multicolumn{2}{c|}{Time Interval (C)}              & \multicolumn{2}{c|}{Time Instant}                   & \multicolumn{2}{c}{No Time}                   \\ \hline
Methods                      & TIMEPLEX  & \multicolumn{1}{l|}{TIME2BOX}       & TIMEPLEX & \multicolumn{1}{l|}{TIME2BOX}       & TIMEPLEX & \multicolumn{1}{l|}{TIME2BOX}       & TIMEPLEX         & TIME2BOX              \\ \hline
MRR                          & \textbf{22.63} & \multicolumn{1}{l|}{22.43}          & 17.72         & \multicolumn{1}{l|}{\textbf{18.85}} & 20.81         & \multicolumn{1}{l|}{\textbf{21.32}} & 67.85                 & \textbf{68.40}        \\
MR                           & 346            & \multicolumn{1}{l|}{\textbf{168}}   & 155           & \multicolumn{1}{l|}{\textbf{147}}   & \textbf{176}  & \multicolumn{1}{l|}{\textbf{193}}   & 430                   & \textbf{172}          \\
HITS@1                       & 4.98           & \multicolumn{1}{l|}{\textbf{11.21}} & 3.94          & \multicolumn{1}{l|}{\textbf{8.35}}  & 11.07         & \multicolumn{1}{l|}{\textbf{11.16}} & \textbf{61.52}        & 60.30                 \\ \hline
\end{tabular}
}
\caption{Link prediction evaluation by types of validity information. Time Interval (O) denotes left/right-open interval-based statements, and Time Interval (C) refers to closed interval-based statements.}
\label{tb:eval_time_prediction_bytype}

\end{table}

\section{Time Prediction Performance by duration length}
Table~\ref{tb:time_prediction_duration_12k} and~\ref{tb:time_prediction_duration_114k} compare the performance of TIMEPLEX and TIME2BOX on the time prediction task across different duration lengths on two datasets. Test samples are first classified into three groups by duration (du) and then evaluate the performance of each group. For an interval $I$,  $du = I_{max}-I_{min}+1$. It shows that our improvements are more pronounced in terms of shorter durations in general. 
\begin{table}[!htbp]
\resizebox{0.5\textwidth}{!}{\setlength\tabcolsep{1.5pt}\begin{tabular}{|l|l|l|l|l|l|l|}
\hline
                    & \multicolumn{6}{c|}{WIKIDATA12k}                                                                                         \\ \hline
Duration (du) & \multicolumn{2}{c|}{du=1}      & \multicolumn{2}{c|}{1\textless{}du\textless{}=5} & \multicolumn{2}{c|}{du\textgreater{}5} \\ \hline
Method              & TIMEPLEX base & TIME2BOX       & TIMEPLEX base         & TIME2BOX               & TIMEPLEX base      & TIME2BOX       \\ \hline
gIOU@1              & 30.29         & \textbf{38.09} & 39.51                  & \textbf{43.68}         & \textbf{47.4}        & 46.99          \\ \hline
aeIOU@1             & 20.84         & \textbf{28.34} & 15.86                  & \textbf{22.95}         & \textbf{18.23}       & 13.20          \\ \hline
gaeIOU@1            & 12.47         & \textbf{18.62} & 11.73                  & \textbf{16.34}         & \textbf{16.85}       & 11.20          \\ \hline
\end{tabular}}
\caption{Time prediction by duration on WIKIDATA12k}
\label{tb:time_prediction_duration_12k}
\end{table}
\begin{table}[!htbp]
\resizebox{0.5\textwidth}{!}{\setlength\tabcolsep{1.5pt}\begin{tabular}{|l|l|l|l|l|l|l|}
\hline
                    & \multicolumn{6}{c|}{WIKIDATA114k}                                                                                        \\ \hline
Duration (du) & \multicolumn{2}{c|}{du=1}       & \multicolumn{2}{c|}{1\textless{}du\textless{}=5} & \multicolumn{2}{c|}{du\textgreater{}5} \\ \hline
Method              & TIMEPLEX base & TIME2BOX       & TIMEPLEX base        & TIME2BOX               & TIMEPLEX  base   & TIME2BOX          \\ \hline
gIOU@1              & 28.75         & \textbf{37.03} & 29.77                  & \textbf{38.36}         & 27.99             & \textbf{39.07}    \\ \hline
aeIOU@1             & 25.80         & \textbf{34.16} & 16.52                  & \textbf{21.54}         & 7.09              & \textbf{9.94}     \\ \hline
gaeIOU@1            & 14.69         & \textbf{21.08} & 10.50                  & \textbf{14.50}         & 3.85              & \textbf{7.02}     \\ \hline
\end{tabular}}
\caption{Time prediction by duration on WIKIDATA114k}
\label{tb:time_prediction_duration_114k}
\end{table}

\section{Model Parameter Comparison}
Table~\ref{tb:num_model_params} summarizes the number of parameters used in each method.
\begin{table}[H]
\centering
\begin{tabular}{|l|l|}
\hline
\textbf{Models} & \textbf{Number of parameters}                  \\ \hline
TNTComplex      & 2d($|E|$ + $|T|$ + 4$|R|$)                           \\ \hline
TIMEPLEX base      & 2d($|E|$ + $|T|$ + 6$|R|$)                           \\ \hline
TIME2BOX        & d($|E|$ + 2$|T|$ + 2$|R|$) + $4d^2$ \\ \hline
\end{tabular}
\caption{Number of parameters for each model}
\label{tb:num_model_params}
\end{table}

\end{appendix}
\end{document}